\documentclass[journal]{IEEEtran}
\usepackage{graphics}
\usepackage{graphicx}
\usepackage{color}
\usepackage{subfigure}
\usepackage{multicol}
\usepackage{multirow}
\usepackage{amssymb}
\usepackage{amsfonts}
\usepackage{float}
\usepackage{booktabs}
\usepackage{psfig}
\usepackage{amsmath}
\usepackage{amssymb}
\usepackage{mathrsfs}
\usepackage{algpseudocode}
\usepackage{epstopdf}
\usepackage{bm}
\usepackage{booktabs}
\usepackage{hyperref}
\usepackage{cite}

\begin{document}

\title{Bi-directional Dermoscopic Feature Learning and Multi-scale Consistent Decision Fusion for Skin Lesion Segmentation}
\author{Xiaohong~Wang, ~Xudong~Jiang, Henghui~Ding, and Jun~Liu\\
Nanyang Technological University, Singapore\\
{\tt\small \{e150023, exdjiang, ding0093, jliu029\}@ntu.edu.sg}
}

\maketitle

\begin{abstract}

Accurate segmentation of skin lesion from dermoscopic images is a crucial part of computer-aided diagnosis of melanoma. It is challenging due to the fact that dermoscopic images from different patients have non-negligible lesion variation, which causes difficulties in anatomical structure learning and consistent skin lesion delineation. In this paper, we propose a novel bi-directional dermoscopic feature learning (biDFL) framework to model the complex correlation between skin lesions and their informative context. By controlling feature information passing through two complementary directions, a substantially rich and discriminative feature representation is achieved. Specifically, we place biDFL module on the top of a CNN network to enhance high-level parsing performance. Furthermore, we propose a multi-scale consistent decision fusion (mCDF) that is capable of selectively focusing on the informative decisions generated from multiple classification layers. By analysis of the consistency of the decision at each position, mCDF automatically adjusts the reliability of decisions and thus allows a more insightful skin lesion delineation. The comprehensive experimental results show the effectiveness of the proposed method on skin lesion segmentation, achieving state-of-the-art performance consistently on two publicly available dermoscopic image databases.

\end{abstract}

\IEEEpeerreviewmaketitle

\section{Introduction}

Melanoma is among the most lethal type of skin cancer that increases rapidly throughout the world, with the five-year survival rate less than 15\% for advanced-stage melanoma~\cite{ma2013analysis,garcia2018segmentation}. Mortality rates of melanoma are associated with its high possibility of metastasizing to other organs (e.g. lung and brain) in the human body~\cite{barata2017development}. Early melanoma usually starts as a brown or black spot that is confined to the cells in the top layer of the skin (epidermis). Then they progressively extend through the epidermis and further into the dermis, followed with the invasion to other tissues and organs through the circulatory system. Timely detection and proper treatment are crucial for patient survival since melanoma can be cured with prompt excision~\cite{xie2017melanoma}. Dermoscopy, widely used in the clinical examination of melanoma, is a noninvasive imaging tool that provides an accurate detail visualization of the pigmented skin lesion structure~\cite{kharazmi2017automated,shimizu2015four}. But manual visual inspection based on dermoscopy is a time-consuming, hardly reproducible and subjective work. Even experienced dermatologists may produce the inconsistent diagnosis results~\cite{yu2017automated}. During recent decades, computer-aided diagnosis systems (CADs) have been developed and already demonstrated strengths for assisting dermatologists in enhancing their clinical diagnosis of melanoma~\cite{cavalcanti2011automated,ahn2017saliency}.

\begin{figure}
\hspace*{-0.6cm}
\vspace*{-0.3cm}
\includegraphics[width=9.6 cm,height=5.1cm ]{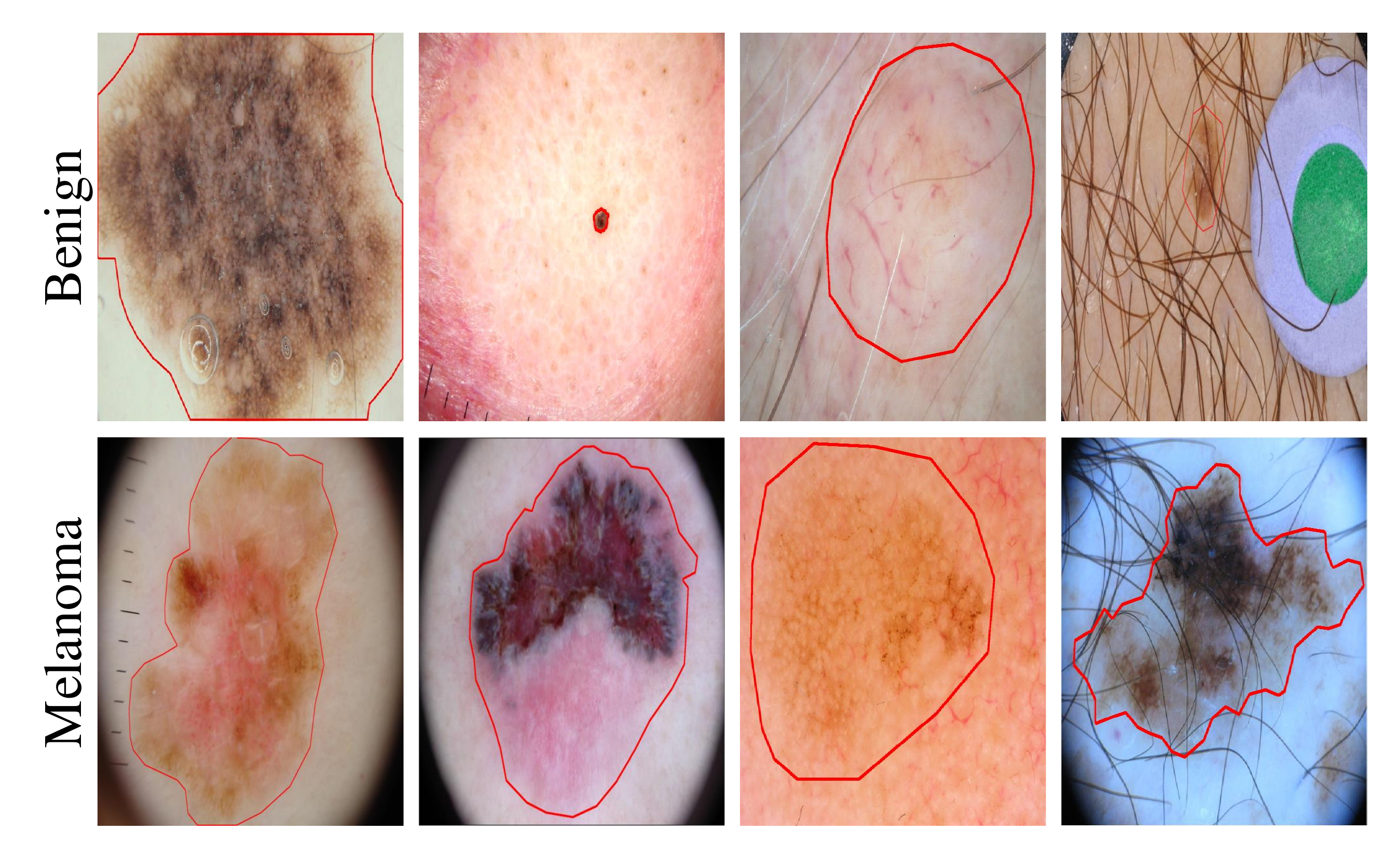}
\caption{Examples of skin lesions in dermoscopic image databases. First and second columns: skin lesions with different shape, size and color. Third column: skin lesions with low contrast compared to background. Fourth column: skin lesions with artifacts.}
\label{fig:1}
\end{figure}

Automatic segmentation of skin lesion is a fundamental component of CADs for the analysis of melanoma~\cite{diaz2018dermaknet}. Recently, fully convolutional neural networks (FCNs)~\cite{yuan2017automatic,bi2017dermoscopic,bi2019step} have shown promising achievements in skin lesion segmentation. The success of FCNs relies on their powerful feature representation ability that encodes both low-level appearance information and high-level semantic information. However, a series of pooling and down-sampling operations at consecutive layers of FCNs reduce the spatial resolution of feature maps and thus yield insufficient skin lesion prediction. Some small skin lesions are so inconspicuous compared to background that FCNs fail to extract valid feature information, though they are important in the diagnosis of melanoma. In addition, the loss of the detailed features in FCNs also limits the localization of skin lesion boundary. Atrous convolutional neural network has displayed its strength in semantic segmentation by handling the feature resolution reduction with multiple parallel dilated convolutional layers~\cite{chen2018deeplab, yang2018denseaspp}. Nevertheless, it is challenging for atrous convolutional neural network to learn a discriminative feature representation for skin lesion due to the non-negligible heterogeneous characteristic of lesions. For instance, there are large appearance variations like shape, size, and color (Fig. \ref{fig:1} (a-b)) of the skin lesion during different lesion progressing stages. Other factors, such as the low contrast between lesion and background, and the presence of artifacts (hairs, air bubbles, color calibration charts, etc), also impede the accurate skin lesion segmentation, as shown in Fig. \ref{fig:1}.

Skin lesions develop as a result of proliferation of a single or multiple components of the skin, which ranges from benign lesions to premalignant lesions and aggressive tumors~\cite{khandpur2012skin}. Since skin lesions progressively invade nearby tissues, there exists a complex correlation between different parts of lesion anatomical structure. 
Utilizing this correlation relationship between candidate pixels and their surrounding contextual regions is beneficial for the network to learn a discriminative feature representation. In this paper, we propose a bi-directional dermoscopic feature learning (biDFL) framework that integrates lesion with their informative context to achieve a substantially rich description of lesion structure. Different from the naive way of FCNs that learns an abstract feature representation for each candidate pixel, our proposed biDFL enriches feature representation by controlling information propagation from two complementary directions among high-level parsing layers. With the integration of both directional feature information passing, the proposed biDFL module enhances the network capability of learning the complex structure of the skin lesion. Without changing the spatial resolution of feature maps, the proposed bi-directional dermoscopic feature learning framework improves the representative capability of feature maps. In addition, our biDFL module also mitigates the challenge of the existence of high variation of skin lesion size.

Different score maps can be generated by classifying features learned from different layers of a neural network, showing the classification results from different scales of learned features. Score maps integration via sum fusion have been adopted to aggregate multi-scale feature information for the refinement of the semantic segmentation in many deep learning based approaches~\cite{long2015fully,shuai2018scene,ding2020semantictip,shuai2019toward,ding2019boundary,wang2019dermoscopic}. However, some skin lesions have large scale while others have small ones. Moreover, pixels far away from the lesion boundary will be more reliably classified by using larger scales of features while pixels near the lesion boundary need to use smaller scale of features for better localization of the segmented lesion. These motivate us to explore the consistency of the classification scores of the local neighborhood for scale selection. Specifically, we propose a multi-scale consistent decision fusion (mCDF) that assesses the reliability of each decision in score maps generated from multiple classification layers. Our multi-scale consistent decision fusion embeds the consistency information around local decision context to adjust the confidence of decision and thus allows more reliable and precise skin lesion delineation. If the decision for a candidate pixel is consistent to its local context decision, then the network gives high confidence for this decision; otherwise the network reduces the confidence of this decision.

Our segmentation network is fully convolutional that provides an end-to-end way for skin lesion training and prediction. Main contributions of this paper are summarized as follows:

$\bullet$ We propose a bi-directional dermoscopic feature learning framework to generate a substantially rich and discriminative feature representation by integrating skin lesions with their informative context. By manipulating the feature propagation through two complementary directions among high-level layers, we improve the image parsing ability of the network.

$\bullet$ We further propose a multi-scale consistent decision fusion to enhance the reliability and consistency of the decision by selectively fusing decisions generated from multiple classification layers.


$\bullet$ We achieve state-of-the-art performance consistently on the evaluated benchmark databases. Even for challenging dermoscopic images, our proposed network also yields high performance on lesion segmentation.


\section{Previous work}

In the past decades, many methods have been reported to deal with the challenges in skin lesion segmentation. Those algorithms can be broadly divided into two categories:  unsupervised and supervised methods.

Unsupervised methods mainly focus on thresholding~\cite{yuksel2009accurate,emre2013lesion}, clustering~\cite{schmid1999segmentation, zhou2009anisotropic}, and deformable contour model~\cite{zhou2011gradient, ma2016novel}. Specifically, Y$\ddot{\rm u}$ksel and Borlu~\cite{yuksel2009accurate} utilized the type-2 fuzzy logic technique for automatic threshold determination. Celebi \emph {et al}.~\cite{yuksel2009accurate} fused four thresholding methods for skin lesion boundary detection. Both thresholding methods separate skin lesions from background based on the histogram distribution of image intensity, which produce undesirable errors for inhomogeneous skin lesions. In~\cite{schmid1999segmentation} and~\cite{zhou2009anisotropic}, fuzzy c-means clustering was employed to segment skin lesion from dermoscopy images. Clustering algorithm assigns pixels with similar characteristics into one identical class, and thus has a limitation for the detection of artificial noises like some hairs and air bubbles. Zhou \emph {et al}.~\cite{zhou2011gradient} segmented skin lesion by using a mean shift based gradient
vector flow (GVF) algorithm. Ma and Tavares~\cite{ma2016novel} exploited a geometric deformable model to simulate the process of the skin lesion segmentation. Deformable models based methods detect skin lesion boundary by minimizing the energy function defined within an image domain. It is difficult for the deformable model to converge around the skin lesions that have low contrast compared to background.

\begin{figure*}
\hspace*{-0.5cm}
\vspace*{-0.3cm}
\includegraphics[width=19 cm,height=10.7cm ]{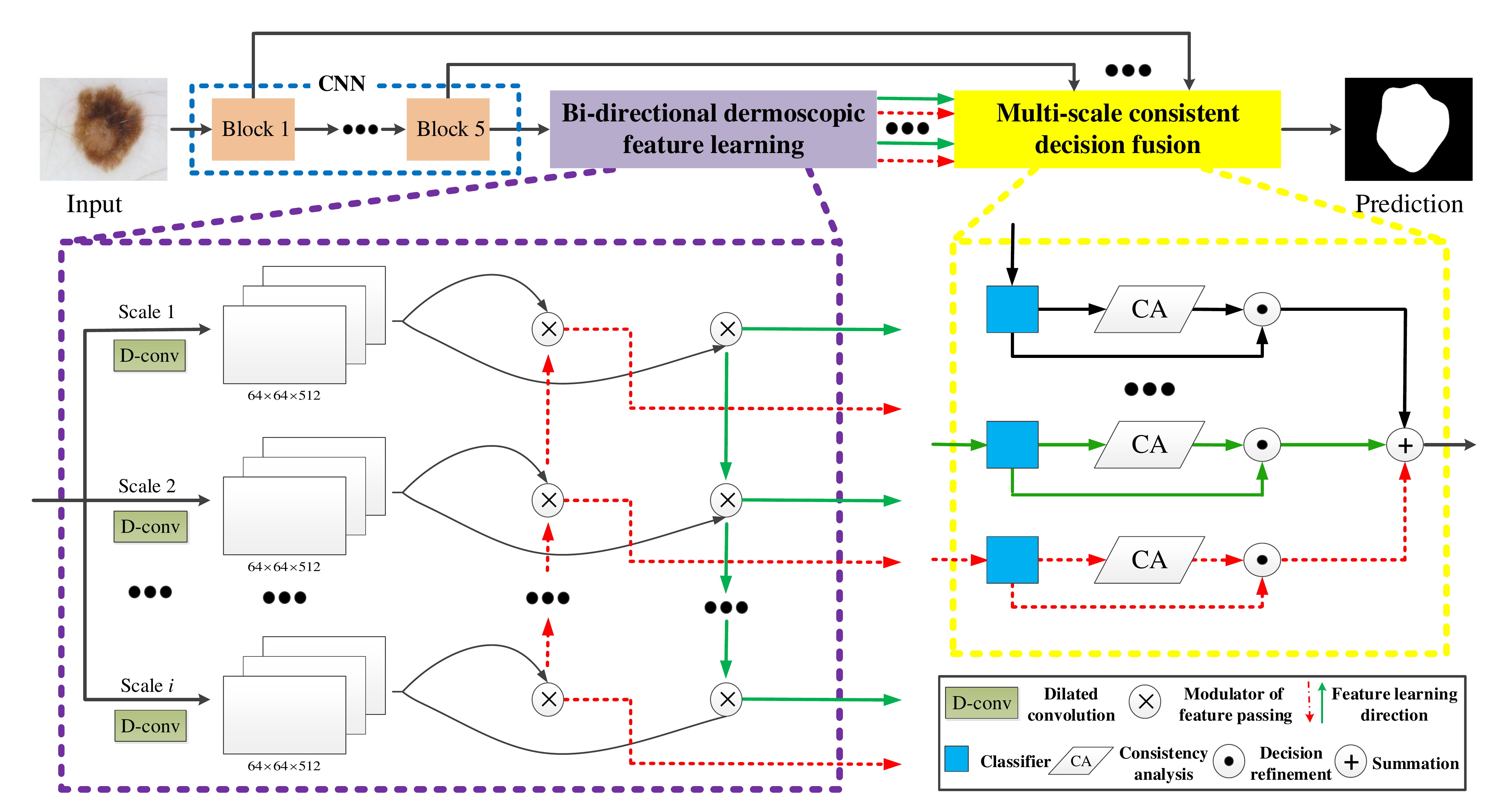}
\caption{Pipeline of the proposed architecture for skin lesion segmentation.  Bi-directional dermoscopic feature learning framework generates a rich lesion representation by the information propagation between the detailed and contextual regions in both directions. Modulator of information passing controls the feature flow between different neurons of network. Green solid arrow denotes the forward feature learning that passes information from local to zooming-out region, while red dot arrow represents the backward feature learning that propagates information from global to zooming-in region. Multi-scale consistent decision fusion embeds consistency analysis in multi-scale decision fusion to achieve reliable and accurate skin lesion prediction. For Blocks 1 to 5, the dimensions of feature maps after each block are $512\times$512$\times$64, $256\times$256$\times$128, 128$\times$128$\times$256, 64$\times$64$\times$512, 64$\times$64$\times$1024 and 64$\times$64$\times$2048 respectively. }
\label{fig:2}
\end{figure*}

Supervised methods can be further grouped into the shallow and deep learning based methods. Shallow learning based methods usually employ hand-crafted features for segmentation, and thus require necessary domain knowledge. For instance, Wang \emph {et al}.~\cite{wang2011modified} deployed 20 descriptive features for lesion segmentation with a neural network classifier. In~\cite{he2012automatic}, a 74-dimension texture feature vector was extracted and then fed into the support vector machine for skin lesion segmentation. Jahanifar \emph {et al}.~\cite{jahanifar2018supervised} applied a supervised saliency detection method tailored for dermoscopic images based on the discriminative regional feature integration. Shallow learning based methods typically depend on capturing appropriate low-level appearance information (e.g. color and texture structure from shallow layer) without combining high-level semantic information, thereby limiting the capability of skin lesion localization as well as the generalization of approaches on other medical image segmentation tasks.


Recently, deep learning achieves great success~\cite{mei2019deepdeblur, liu2019feature} and deep learning based segmentation techniques have been reported in skin lesion prediction following their success in other medial image analysis fields, such as multi-modal brain tumor segmentation~\cite{pereira2016brain, wang2018interactive}, gland segmentation~\cite{chen2016dcan}, pulmonary nodule detection~\cite{setio2016pulmonary}, and body organs recognition~\cite{yan2016multi}. Gu \emph{et al}. integrated the dense atrous convolution module and residual multi-kernel pooling with encoder-decoder structure for the segmentation of optic disc, retinal vessel, lung, cell contour and OCT layer~\cite{gu2019net}. Zhang \emph{et al}. embedded edge-attention representations to guide the process of segmentation on optic disc, retinal vessel, and lung~\cite{zhang2019net}. Attention modules incorporated in deep learning architectures have also shown their strengths in many computer vision based tasks~\cite{woo2018cbam,hu2018squeeze}. Schlemper \emph{et al}. encapsulated attention gates into a 3D U-Net architecture for abdominal organ segmentation~\cite{schlemper2019attention}. The target of attention gates is to highlight salient features that are passed through the skip connections. Wang \emph{et al}. built a 3D attention guided deep learning network for prostate segmentation by harnessing the spatial context across deep and shallow layers~\cite{wang2019deep}. It refines the features at each individual layer by selectively leveraging the multi-level features integrated from different layers. For skin lesion segmentation task, Yuan \emph {et al}.~\cite{yuan2017automatic} segmented skin lesion using deep fully convolutional networks (FCNs) with Jaccard distance. Yu \emph {et al}.~\cite{yu2017automated} utilized a fully convolutional residual network incorporating a multi-scale contextual information integration scheme to automatically segment skin lesion. Bi \emph {et al}.~\cite{bi2017dermoscopic} presented a multi-stage segmentation approach where early-stage FCNs extracted coarse appearance information and late-stage FCNs learned the subtle characteristics of the lesion boundaries. Yuan \emph {et al}.~\cite{yuan2017improving} designed a deeper network architecture with more smaller convolutional kernels than their previous work~\cite{yuan2017automatic}, and investigated the efficiency of using the channels from Hue-Saturation-Value color space. In~\cite{lin2017skin}, the authors applied U-net for skin lesion segmentation. Al-masni \emph {et al}.~\cite{al2018skin} achieved pixel-wise segmentation of the skin lesion by a full resolution convolutional networks (FrCn), which eliminated the subsampling layers in the networks and enabled the convolutional layers to extract and learn the full spatial features of the input image. 
Bi \emph {et al}.~\cite{bi2019step} trained the deep ResNet model~\cite{he2016deep} independently across different classes and refined the segmentation performance by a probability based step-wise integration.

Different from the aforementioned medical image segmentation approaches, our proposed network investigates the complex correlation between skin lesions and their informative context to achieve a discriminative feature representation, which leads to more robust high-level parsing. Furthermore, a multi-scale consistent decision fusion technique is proposed to make more reliable and precise skin lesion prediction by analyzing the consistency of decisions in a local area. Quantitative and qualitative evaluations show the superiority of the proposed network on skin lesion segmentation.

\section{Proposed Framework}

In this paper, we tackle the challenging task of skin lesion segmentation. Fig. \ref{fig:2} shows the overall framework of the proposed network, where a FCN-like architecture with ResNet50 (pre-trained on ImageNet~\cite{russakovsky2015imagenet}) is applied as our baseline network. On top of it, we control feature information passing with a series of modulators from two directions: the forward direction from local to zooming-out region (marked as green solid arrow in Fig. \ref{fig:2}) and the backward direction from global to zooming-in region (marked as red dot arrow in Fig. \ref{fig:2}). The forward feature propagation simulates the human visual perceptual process that cerebral cortex encoding visual stimulus starts from extracting local feature of image in the lower visual pathways, followed by integrating the local features into global features in the higher visual pathways~\cite{hubel1995eye}. This integration of local features into more global ones dominates the field of the anatomical, physiological and behavioral studies of vision system~\cite{huang2017rapid}. The other direction of information passing follows another visual perceptual mechanism that relies on the feature cognition from global to local~\cite{ahissar2004reverse,zoccolan2015invariant}. With this bi-directional dermoscopic feature learning design, local and contextual features cooperate with each other on the skin lesion description. Furthermore, multi-scale consistent decision fusion (shown in the yellow box in Fig. 2) helps the network selectively combine the informative decisions from multiple classification layers, which leads to the improvement of the reliability and consistency of the predication. Details of the proposed bi-directional dermoscopic feature learning and multi-scale consistent decision fusion are described in the following sections.




\subsection{Context Feature Map Generation with Multiple Dilation Rates}

Deep fully convolutional neural network based methods~\cite{bi2019step,long2015fully}, progressively reduce the feature map resolution by a series of consecutive pooling and convolution striding operations. This produces the abstract feature representation that characterizes local areas of an image. Different from the object recognition that classifies the whole input image, the segmentation task requires classifying every local pixel separately. Thus, we need to generate high-level semantic context features for every local pixel for more reliable classification. To recognize the complex structure of the skin lesion, it is essential to form a discriminative high-level semantic context feature representation for skin lesion in dermoscopic image. Since the semantic meaning of feature maps generated at different spatial configurations of the skin lesion is correlated, information passing through each part of the lesion structure can effectively improve the feature description of the skin lesion. Herein, we explore an effective way to integrate the information of skin lesions with their context.

Let $F_{0}$ denote the feature maps generated at the top layer of a pretrained CNN, i.e. the output of the Block 5 as illustrated in Fig. \ref{fig:2}. The multi-scale context feature maps are produced by a series of dilated convolutions with ascending dilation rates $\gamma_j$:
\begin{equation}
F_{j} = \mathscr{F}_{\gamma_{j}}(F_{0},\Theta_{\textit{j}}), j\in[1,2,..,J],
\end{equation}

\noindent where $\mathscr{F}_{\gamma_{j}}$ is the function of 3$\times$3 convolution with the dilation rate $\gamma_{j}$, $\Theta_{\textit{j}}$ are the respective parameters, and $J$ is the number of dilation rates. The set of feature maps $F_{j}(j\in[1,2,..,J])$ cover a rich context information of skin lesion since $J$ different sizes of the receptive field of dilated convolution filters are applied to perceive the information near and far. The detailed relationship of different dilation rates $\gamma_{j}$ and their corresponding capability of encoding spatial contextual information are concretely elaborated as

%


(1) $Local$: Dilated convolutions with small rates focus on the extraction of features around the local region, which allows the network to form a detailed feature representation. Small dilation rates preserve locality of features but are sensitive to artificial signals emerged during dermoscopic image acquisition, like noises and air bubbles. Small dilation rates provide more accurate boundary information but have a high potential to lead the network to predict many isolated fake lesion tissues.


(2) $Regional$:  With the increasing dilation rates, the receptive fields of dilated convolution filters are enlarged to capture much larger spatial contextual information of skin lesions. Compared to small dilation rates, those medium dilation rates allow the network to harness larger scale of contextual information more efficiently and thus generate prediction more robust to the challenging artificial noises, irregular lesion tissues, as well as some inhomogeneous dermoscopic images. Since medium dilation rates are not fit for matching the size of extremely tiny or large skin lesions, features extracted with medium dilation rates are hard to characterize both of them.


(3) $Global$: Large dilation rates further enlarge the receptive fields of dilated convolution filters that are suitable to capture skin lesion of large scales. Dilated convolutions with large dilation rates are able to see more global skin lesion pattern, but reduce the power of capturing detailed local information and thus yield coarse prediction for skin lesion.


\begin{figure}
\hspace*{-0.4cm}
\vspace*{-0.3cm}
\includegraphics[width=9.3 cm,height=6.1cm ]{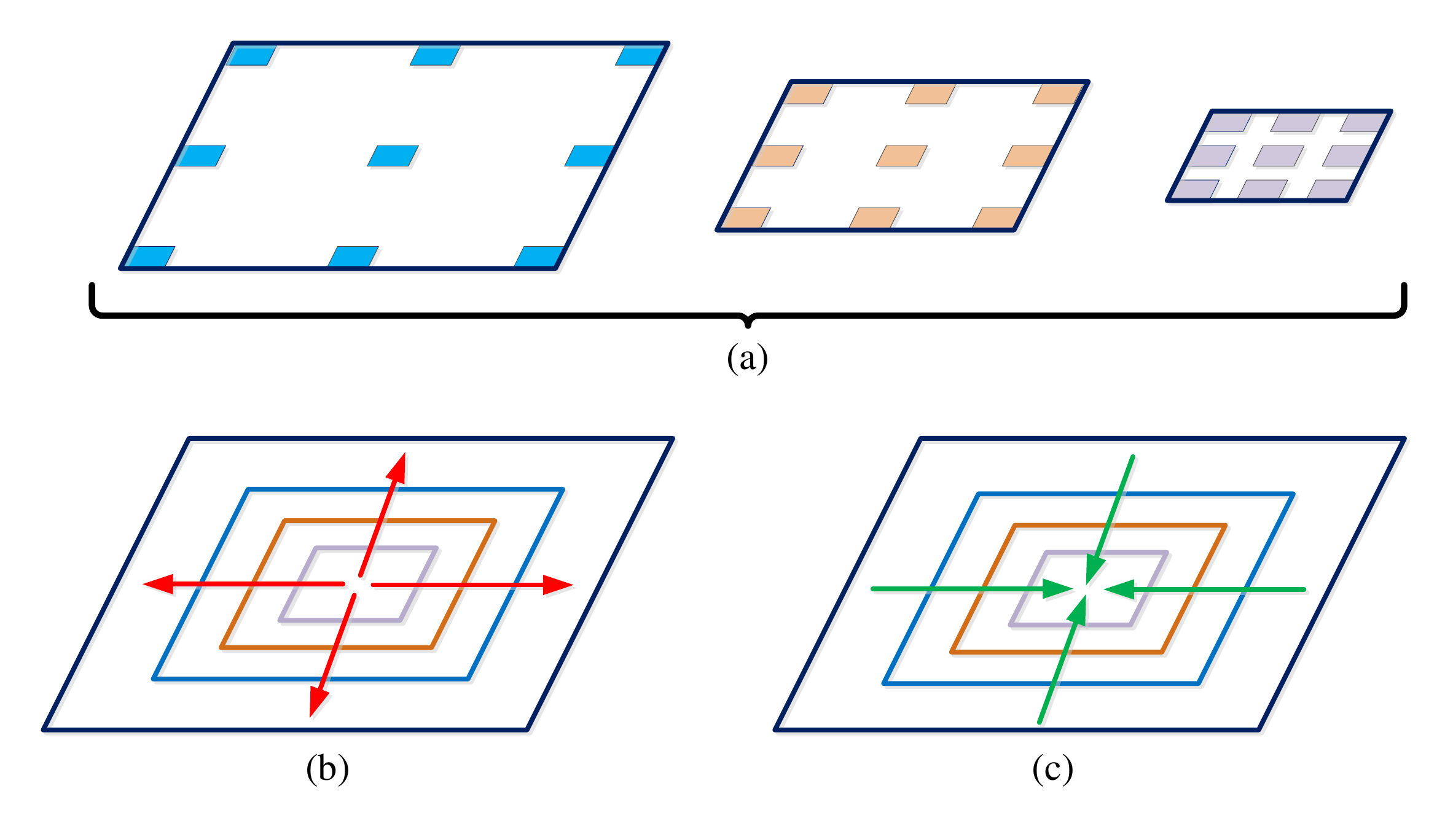}
\caption{Bi-directional dermoscopic feature learning process. (a) Receptive fields of dilated convolution filters with three different rates. (b) Feature propagation from local to zooming-out region. (c) Feature propagation from global to zooming-in region.}
\label{fig:3}
\end{figure}

\subsection{Bi-directional Dermoscopic Feature Learning}

A common way of utilizing the feature maps generated by dilated convolutions with multi-scale rates is to concatenate them directly. For the task of skin lesion segmentation, simple concatenation largely increases the feature dimensionality that will degrade the generalization capability of the feature maps for classification. In this paper, we propose a bi-directional dermoscopic feature learning module to perceive the information of lesion characteristics. It controls the propagation of the feature flow among the complex spatial configuration of the skin lesion. Fig. \ref{fig:3} illustrates the process of the feature information passing along two directions. Given three feature maps obtained from dilated convolutions with different rates as shown in Fig. 3 (a), the complex correlation between skin lesion and their informative context can be captured through two directions: one from local to zooming-out region (Fig. \ref{fig:3} (b)) and the other from global to zooming-in region (Fig. \ref{fig:3} (c)), respectively. By integrating both directional feature learning propagation, the proposed biDFL module allows the network to form a substantially rich description of skin lesion.

More specifically, the feature maps $F_{j}$ after information propagation from local to zooming-out region are refined to be

\begin{equation}
F^{'}_{j} = \mathscr{F}(\mathcal{C}(F^{'}_{j-1};F_{j}),\hat{\Theta}_{\textit{j}}), j>1, j\in[1,2,..,J],
\label{eq:2}
\end{equation}

\noindent where $\mathcal{C}$ is the concatenation process, $\mathscr{F}$ is the function of $1\times1$ convolution, and $\hat{\Theta}_{\textit{j}}$ are the respective parameters. The number of channels of $F^{'}_{j}$ is kept the same as that of $F_{j}$ through a number of 1$\times$1 convolutions. This progressively accumulates the informative context $F_{1}$,..., $F_{j-1}$, $F_{j}$ into $F^{'}_{j}$ that will have more powerful representation for the complex structure of skin lesion. Especially for $F_{j}$ generated from a large receptive field, the degradation of representation ability of feature maps $F_{j}$ generated by simple dilated 3$\times$3 convolutions are alleviated by the proposed feature information propagation.

To obtain the complementary and detailed features for skin lesion, we exploit another information passing that has the opposite propagation direction. The refined feature maps $F^{''}_{j}$ through this direction are updated as

\begin{equation}
F^{''}_{j} = \mathscr{F}(\mathcal{C}(F^{''}_{j+1};F_{j}),\tilde{\Theta}_{\textit{j}}), j<J, j\in[1,2,..,J].
\label{eq:2}
\end{equation}

\noindent As formulated in Eqs (2-3), the correlation between skin lesions and their informative context is learned through a series of modulators ($\mathcal{C}$ and $\mathscr{F}$). The feature maps $F^{'}_{j}$ receive the messages through the forward feature learning process that aggregates a series of zooming-out context feature information from $F_{l}$ ($l<j$). Moreover, the feature maps $F^{''}_{j}$ obtain messages through the backward feature learning process that fuses a set of zooming-in context feature information from $F_{m}$ ($m>j$). By combining the feature propagation along both directions, our proposed biDFL achieves a discriminative feature representation for lesion where information progressively passes among the complex spatial configuration of the skin lesion.

Stacking convolution layers together by two directions also effectively enriches the set of feature maps with multiple receptive fields. This is important for skin lesion segmentation since features for each lesion pixel can be represented more discriminatively by propagating information flow between those consecutive convolution layers of the proposed framework. Especially for some skin lesions with similar appearance as their non-lesion neighborhoods, without the support of information from proper receptive fields, they are hard to be distinguished from background. With the design of sequentially aggregating the features from multiple receptive fields, our proposed biDFL assists the network to learn  feature description with rich multi-scale information, which is helpful for those challenging cases.

\subsection{Multi-scale Consistent Decision Fusion}

%
%

Dermoscopic images have a large variation in size of skin lesions. Let $P$ denote the size of the skin lesion relative to that of the respective dermoscopic image. We found great variation of the $P$ value in the dermoscopic image database. Specifically, the maximum value $P$ for ISBI 2016 and ISBI 2017 databases are 0.9954 and 0.9522 respectively, where the minimum value $P$ for ISBI 2016 and ISBI 2017 databases are 0.0027 and 0.0030 respectively. This shows that there exist relatively tiny and inconspicuous skin lesions compared to background. For traditional networks like VGG net~\cite{simonyan2014very} and Resnet-50~\cite{he2016deep}, the spatial resolutions of the feature maps are usually reduced by a factor of 32 after a series of down-sampling processes. It is hard for those networks to capture the information of the aforementioned skin lesions, though they play an important role in the diagnosis of melanoma. In addition, the boundary of skin lesions are complex curves that are difficult to be precisely delineated by the down-sampled feature maps with low resolution.

To further enhance the prediction performance, we exploit features from shallow layers as they contain more detailed information of the inconspicuous skin lesions as well as complex boundary structure. Previous works~\cite{long2015fully,peng2017large,shuai2018scene,shuai2019toward}, mainly integrate score maps non-selectively from different skip layers, where some inappropriate scores may decrease the precision of the skin lesion prediction. For example, the score maps from low-level layers may contain debatable noisy information in homogeneous region. On the other hand, the scores from high-level layers of pixels located near the boundary of skin lesion are less informative, since they are insensitive to the spatial location of the skin lesion. Therefore, pixels far away from the lesion boundary will be more reliably classified by using larger scales of features in high-level layers while pixels near the lesion boundary need to use smaller scales of features in low-level layers for better localization of the segmentation boundary. An adaptive score map aggregation, incorporating the selection of reliable scale features for each pixel, is beneficial to skin lesion segmentation. In this paper, we propose a multi-scale consistent decision fusion strategy that selectively aggregates score maps by controlling the reliability of multi-scale feature representation among skip layers. With the embedding of consistency analysis to the decisions from each classification layer, our proposed mCDF assists the network to learn better about which scales of features are more desirable for each individual pixel.   

Suppose for a class (here skin lesion or background), there are $K$ score maps $S^{k}_{p}$ generated by $K$ skip layers from different scales of features, where $p$ is the spatial position and $k\in K$. We compute the coefficient of mCDF, $\alpha^{k}_{p}$, by assessing the consistency of the $k$th decision within a local region $L_{p}$ centered at $p$. We formulate $\alpha^{k}_{p}$ as a Gaussian function of the standard deviation of the scores $S^{k}_{p}$ over a spatial local region $L_{p}$:
\begin{equation}
\alpha^{k}_{p} = e^{\frac{-(\sigma_{p}^{k})^2}{\sigma^{2}}},
\end{equation}
\noindent where  \begin{equation}
\sigma_{p}^{k} = \sqrt{\mathop{\rm mean}\limits_{L_{p}} (S_{p}^{k}-\mathop{\rm mean}\limits_{L_{p}} (S_{p}^{k}))^2} .
\end{equation}

\noindent Gaussian function supports the decision consistency varying in an effective range, i.e. from 0 to 1. If the prediction from a skip layer for a pixel is consistent to its context, the output of Gaussian function is close to 1. Otherwise it decays to 0 quickly. In addition, the form of Gaussian function has the benefit of the implementation of the gradient back-propagation. Parameter $\sigma$ controls the sensitivity of the consistency coefficient $\alpha^{k}_{p}$ to the variation of the score  $S^{k}_{p}$ in the local region. For each position $p$, we compute multi-scale consistent coefficients $\alpha^{k}_{p}$ for $K$ skip layers. The coefficient $\alpha^{k}_{p}$ reflects the consistency of decision in a local region centered at position $p$ at the $k$th skip layer. A smaller coefficient $\alpha^{k}_{p}$ implies less reliable of the decision $S^{k}_{p}$ at the $k$th skip layer, so score fusion process should suppress the effect of $S^{k}_{p}$. A larger coefficient $\alpha^{k}_{p}$ means that for position $p$, the decision $S^{k}_{p}$ at the $k$th skip layer is more desirable and has high potential to achieve an accurate prediction. As a consequence, the effect of $S^{k}_{p}$ with a larger coefficient $\alpha^{k,c}_{p}$ should be highlighted during the decision fusion.

Under the control of the consistent coefficient $\alpha^{k}_{p}$, all score maps are selectively fused by
\begin{equation}
S_{p}=\sum\limits^{K}_{k=1}\alpha^{k}_{p}S^{k}_{p}.
\end{equation}
\noindent $\alpha^{k}_{p}$ is adaptive to the score map $S^{k}_{p}$ without complicated heuristic learning process. $\alpha^{k}_{p}S^{k}_{p}$ is the decision refinement mechanism that adjusts the contribution of the decision $S^{k}_{p}$ of each skip layer by the factor $\alpha^{k}_{p}$. In contrast to the simple score summation, the mCDF depends on the consistency of the score map at each skip layer to achieve selectively multiple decisions fusion. Moreover, the model trained with mCDF implicitly learns to inhibit irrelevant noises while adaptively merging rich feature information at multiple scales. Therefore, by adding multi-scale consistent decision fusion, our proposed network is more powerful to recover the details in shallow layers and yield reliable pixel-wise skin lesion prediction maps.

\section{EXPERIMENTAL EVALUATION}

\subsection{Materials}
We test the proposed framework on two publicly available databases (ISBI 2016~\cite{timmurphy16} and ISBI 2017~\cite{timmurphy17}) provided by the International Skin Imaging Collaboration (ISIC) archive. ISBI 2016 and ISBI 2017 are two challenge databases of ``Skin Lesion Analysis toward Melanoma Detection'' hosted by the International Symposium on Biomedical Imaging (ISBI) in 2016 and 2017 respectively.

$\bullet$ ISBI 2016, comprises 900 training images and 379 test images in JPEG format. Ground truths for ISBI2016, created by an expert clinician, are encoded as single-channel (grayscale) 8-bit PNGs. Training and test images are diagnosed as non-melanoma or melanoma, resulting in 727 non-melanoma and 173 melanoma in training set, 304 non-melanoma and 75 melanoma in test set.

$\bullet$ ISBI 2017, the extension of ISBI2016, consists of 2000 training images and 600 test images in JPEG format. Training set contains 374 melanoma, 254 seborrheic keratosis, and 1372 benign nevi, while test set includes 117 melanoma, 90 seborrheic keratosis, and 393 benign nevi. In addition, ISBI 2017 database also provides the validation set that includes 150 skin lesion images.

\subsection{Evaluation Criterion}
Performance of the proposed method is evaluated by comparing the skin lesion segmented result with the ground truth created by each database. Four different measurements for skin lesion segmentation performance evaluation include Jaccard index (JA), Dice coefficient (DI), segmentation accuracy (AC), and G-mean (GM). JA and DI measure the similarity between the detected result and ground truth: JA = $TP/(TP+FN+FP)$ and DI = $2TP/(2TP+FN+FP)$, where $TP$ and $TN$ represent the number of pixels correctly classified as lesion and background pixels; $FN$ and $FP$ denote the number of pixels incorrectly classified as the background and lesion pixels, respectively. AC is the pixel-wise accuracy, i.e. the ratio of correctly detected pixels to total pixels. GM is a metric that estimates the imbalance between segmentation sensitivity and specificity by taking their geometric mean: GM = $\sqrt{TP\times TN/(TP+FN)(TN+FP)}$. ISBI 2016 and 2017 challenges took Jaccard index (JA) as the most important criterion for segmentation comparison and participants were ranked based on it.
%
\begin{table*}[t]
\centering
\setlength{\abovecaptionskip}{0pt}
\setlength{\belowcaptionskip}{100pt}
\caption{SEGMENTATION PERFORMANCE ON ISBI 2016 AND ISBI 2017 DATABASES (mean$\pm$standard deviation,\%)}
\renewcommand\arraystretch{1.4}
\setlength{\tabcolsep}{1.7pt}
\label{tab:1}
\resizebox{\textwidth}{23mm}{
\begin{tabular}{cclllllllllllllll}
\toprule
\toprule
\multicolumn{1}{c}{\multirow{2}{*}{Database}} &\multicolumn{1}{c}{\multirow{2}{*}{Method}} &\multicolumn{4}{c}{Melanoma} &\multicolumn{4}{c}{Non-Melanoma} &\multicolumn{4}{c}{Overall} \\
\cline{3-6} \cline{7-10} \cline{11-14}
& &\multicolumn{1}{c}{JA} &\multicolumn{1}{c}{DI} &\multicolumn{1}{c}{AC} &\multicolumn{1}{c}{GM} &\multicolumn{1}{c}{JA} &\multicolumn{1}{c}{DI} &\multicolumn{1}{c}{AC} &\multicolumn{1}{c}{GM} &\multicolumn{1}{c}{JA} &\multicolumn{1}{c}{DI} &\multicolumn{1}{c}{AC} &\multicolumn{1}{c}{GM}\\
\midrule
&Baseline &85.03$\pm$13.69 &91.56$\pm$11.35 &94.57$\pm$7.33 &93.73$\pm$10.87 &84.33$\pm$9.99 &90.83$\pm$6.48 &95.84$\pm$4.02 &93.99$\pm$6.84 &84.47$\pm$13.04 &90.97$\pm$10.57 &95.59$\pm$6.82 &93.94$\pm$10.19 \\
ISBI &Baseline+biDFL 
&87.24$\pm$10.81 &92.74$\pm$8.09 &95.13$\pm$4.95 &93.96$\pm$5.38 &87.18$\pm$9.28 &92.52$\pm$5.83 &96.48$\pm$3.97 &94.79$\pm$11.82 &87.19$\pm$10.52 &92.56$\pm$7.69 &96.21$\pm$4.79 &94.63$\pm$7.16 \\
2016 &Baseline+mCDF &87.03$\pm$13.12 &92.86$\pm$10.99 &95.10$\pm$7.39 &94.45$\pm$7.82 &85.86$\pm$8.00 &91.64$\pm$4.84 &96.11$\pm$4.04 &94.90$\pm$4.62 &86.09$\pm$12.28 &91.88$\pm$10.08 &95.91$\pm$6.87 &94.82$\pm$7.29 \\
&Baseline+biDFL+mCDF &88.38$\pm$9.99 &93.62$\pm$7.52 &95.95$\pm$4.94 &94.53$\pm$6.49 &88.06$\pm$8.13 &93.26$\pm$4.96 &96.95$\pm$3.23 &95.35$\pm$8.40 &88.12$\pm$9.64 &93.33$\pm$7.08 &96.75$\pm$4.67 &95.19$\pm$6.91 \\
\midrule
&Baseline  &69.24$\pm$20.69 &79.12$\pm$18.75 &90.09$\pm$14.01 &84.22$\pm$18.14 &77.32$\pm$21.64 &85.30$\pm$19.20 &93.87$\pm$13.77 &89.49$\pm$16.45 &75.74$\pm$20.91 &84.10$\pm$18.93 &93.13$\pm$14.02 &88.47$\pm$17.88\\
ISBI 
&Baseline+biDFL &72.91$\pm$18.28 &82.03$\pm$15.04 &91.01$\pm$10.71 &86.05$\pm$15.51 &80.32$\pm$21.44 &87.59$\pm$18.98 &94.74$\pm$10.04 &90.61$\pm$14.53 &78.87$\pm$19.15 &86.51$\pm$16.01 &94.01$\pm$10.68 &89.73$\pm$15.43\\
2017&Baseline+mCDF &73.25$\pm$18.82 &82.88$\pm$16.35 &90.15$\pm$9.48 &87.10$\pm$12.35 &78.20$\pm$18.60 &86.08$\pm$16.13 &93.99$\pm$10.40 &90.97$\pm$12.23 &77.24$\pm$18.87 &85.45$\pm$16.34 &93.24$\pm$9.78 &90.21$\pm$12.42\\
&Baseline+biDFL+mCDF &77.26$\pm$15.83 &85.35$\pm$12.80 &92.02$\pm$9.61 &89.74$\pm$13.4 &82.49$\pm$19.37 &89.31$\pm$17.04 &95.28$\pm$9.34 &93.37$\pm$11.91 &81.47$\pm$16.69 &88.54$\pm$13.80 &94.65$\pm$9.64 &92.67$\pm$13.22 \\
\bottomrule
\bottomrule
\end{tabular}}
\end{table*}

\subsection{Implementation Details}

We employ a FCN-like architecture with ResNet50 (pre-trained on ImageNet~\cite{russakovsky2015imagenet}) as our baseline network. In detail, we first remove the last pooling layer and layers after it. For the last two blocks of the network, we keep the resolution of feature maps unchanged but set the dilation rates of convolution layers at the two blocks to be 2 and 4 respectively, which allows the reuse of the pre-trained weights. For Blocks 1, 2 and 3, the dimensions of feature maps after each block are $256\times256\times64$, $128\times128\times256$, and $64\times64\times512$ respectively. For Blocks 4 and 5, the dimensions of feature maps after each block become $64\times64\times1024$ and $64\times64\times2048$. We generate five sets of feature maps with five different dilation rates for the output of Block 5, i.e. (3, 6, 12, 18, 24), which are the extension of the four dilation rates of ASPP with large FOV~\cite{chen2018deeplab}. To avoid the network growing too wide and reduce the computer consumption, we apply a $1\times1$ convolution layer before the dilated convolution to decrease the number of feature channels from 2048 to 512.

Since the distribution of lesion and non-lesion pixels are unbalanced, i.e. lesion pixels occupy a relatively small proportion on both ISBI 2016 and ISBI 2017 databases, this paper uses the weighted cross entropy loss to measure the difference between the true label and the predicted result, shown as~\cite{shuai2018scene,shuai2019toward,wang2019dermoscopic}:
\begin{equation}
L = -\frac{1}{N}\sum_{p\in I} \sum_{i=1}^{C} w_{i}y_{i}^{p}log(o^{p}_{i}),
\end{equation}
\noindent where $N$ is the number of pixels in image $I$. $y_{i}^{p}$ is the true label for the pixel located at $p$. $o^{p}_{i}$ is the corresponding class likelihood. $C$ is the number of categories. Here $i$ changes from 1 to 2, where 1 indexes lesion class and 2 indexes background class. $w_{i}$ is the weight for class $i$. We set 0.8 for lesion class and 0.2 for background class to alleviate the unbalance problem, since the lesion class has only about 20\% of the pixels in the whole images. Inspired by~\cite{ding2019semantic}, we apply the ``poly" learning rate policy where the current learning rate is multiplied by $(1-\frac{iter}{max_{-}iter})^{power}$. The initial learning rate and power are set to $10^{-3}$ and 0.9 respectively. $\sigma^2$ is empirically set to 10 for controlling the sensitivity of the consistency coefficient to the variation of the score. Stochastic gradient descent (SGD) algorithm is exploited to train our end-to-end network. The number of iterations is set to 30k for ISBI 2017 database and 12k for ISBI 2016 database respectively. To capture the effective feature maps at different scales, we design a controllable local region size $l\times l$ for $L_{p}$. For feature maps from the first three Blocks, $l\times l$ is set to $3\times3$ as each block has a down-sampling process. For feature maps from Blocks 4 to 5, as well as the five dilated convolution layers, $l\times l$ is progressively enlarged as $5\times5$, $7\times7$,...,$17\times17$, respectively.

Skip layers utilize convolution transpose kernels~\cite{ding2018context} to achieve up-sampling operation. For batch processing, all images are resized to have maximum extent of 512 pixels. Since ISBI 2016 database only provides a training set and a test set, we randomly select 800 images from training set for training, and the rest 100 images in training set for validation. For ISBI 2017 database, we conduct training, validation, and testing on its provided training, validation and test sets, respectively. One of the challenges of skin lesion segmentation is the insufficiency of the training data with high quality. Augmentation strategy including random flipping images (horizontally, vertically), and random scaling in the range of [0.8 , 1.2], are performed to generate more diverse training data.

\begin{figure*}
\hspace*{-0.4cm}
\includegraphics[width=18.7 cm,height=8.3cm ]{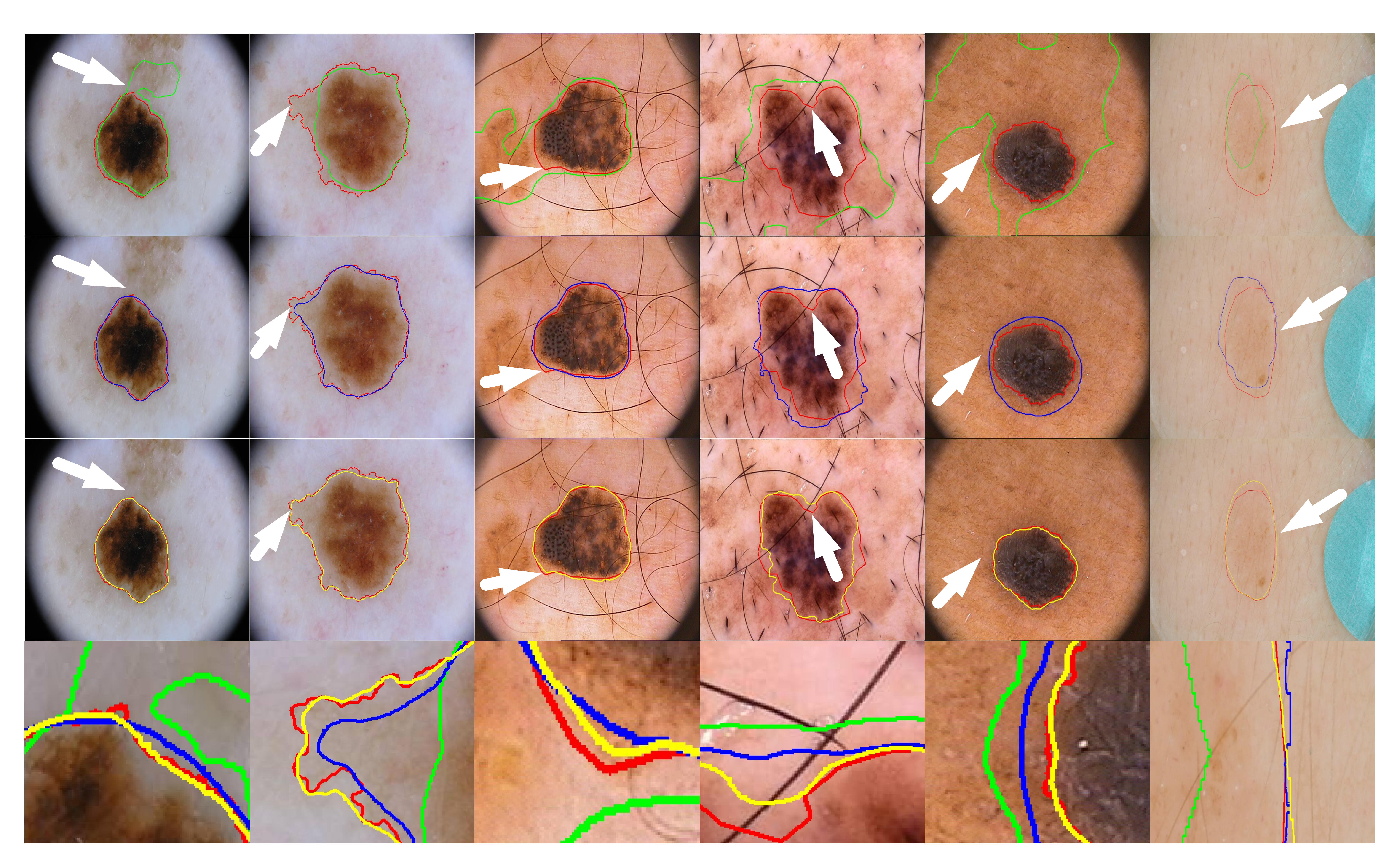}
\vspace*{-0.5cm}
\caption{Qualitative segmented result comparison on ISBI 2016 database. First row to third row: segmentation results by baseline (green lines), baseline +biDFL (blue lines) and baseline+biDFL+mCDF (yellow lines) respectively, where ground truths are annotated as red lines. Last row: detailed segmentation results of sub-regions (marked by white arrows in the corresponding columns).}
\label{fig:4}
\end{figure*}

\begin{table*}
\centering
\hspace*{0.6cm}
\setlength{\abovecaptionskip}{0pt}
\caption{Component investigation of bi-directional dermoscopic feature learning (mean$\pm$standard deviation,\%)}
\renewcommand\arraystretch{1.3}
\setlength{\tabcolsep}{5pt}
\label{tab:2}
\begin{tabular}{lllllllll}
\toprule
\toprule
\multicolumn{1}{l}{\multirow{2}{*}{Method}} &\multicolumn{4}{c}{ISBI 2016} &\multicolumn{4}{c}{ISBI 2017} \\
\cline{2-5} \cline{6-9}
 &\multicolumn{1}{c}{JA} &\multicolumn{1}{c}{DI} &\multicolumn{1}{c}{AC} &\multicolumn{1}{c}{GM} &\multicolumn{1}{c}{JA} &\multicolumn{1}{c}{DI} &\multicolumn{1}{c}{AC} &\multicolumn{1}{c}{GM} \\
\midrule
Baseline &84.47$\pm$13.04 &90.97$\pm$10.57 &95.59$\pm$6.82 &93.94$\pm$10.19 &75.74$\pm$20.91 &84.10$\pm$18.93 &93.13$\pm$14.02 &88.47$\pm$17.88\\
Baseline+${\rm DFL}_{+}$ &86.01$\pm$11.00 &91.99$\pm$8.33 &95.83$\pm$6.35 &93.82$\pm$8.61 &78.37$\pm$19.33 &86.21$\pm$16.09 &93.58$\pm$11.40 &89.58$\pm$16.27\\
Baseline+${\rm DFL}_{-}$ &85.92$\pm$11.45 &91.93$\pm$8.26 &96.16$\pm$5.03 &94.03$\pm$7.82 &78.03$\pm$19.76 &85.98$\pm$16.55 &93.83$\pm$11.91 &89.76$\pm$16.55\\
Baseline+biDFL &87.19$\pm$10.52 &92.56$\pm$7.69 &96.21$\pm$4.79 &94.63$\pm$7.16 &78.87$\pm$19.15 &86.51$\pm$16.01 &94.01$\pm$10.68 &89.73$\pm$15.43\\
\bottomrule
\bottomrule
\end{tabular}
\end{table*}
\subsection{Ablation Studies}

\subsubsection{Evaluation of individual component in our approach}
To evaluate the proposed architecture on skin lesion segmentation, we conduct step-by-step ablation experiment on dermoscopic images in ISBI 2016 and ISBI 2017 databases. The detailed quantitative experimental results are shown in Table \ref{tab:1}, where we can observe that each proposed contribution collectively improves the baseline network on skin lesion segmentation. By jointly learning feature from the complex spatial configuration of skin lesion through two complementary directions, we improve the segmentation performance JA for ISBI 2016 and ISBI 2017 databases by 2.7\% and 3.1\% respectively. By integrating the multi-scale consistent decision fusion, we enhance the segmentation performance JA on both databases about 1.5\%-1.6\%. Combination of bi-directional dermoscopic feature learning and multi-scale consistent decision fusion achieves a performance superior to the baseline network, with the JA enhancement of 3.7\%-5.7\% on both databases. In addition, the significant segmentation performance improvement consistently on melanoma and non-melanoma cases in both databases, demonstrates the reliability and robustness of the proposed architecture on dealing with the task of skin lesion segmentation.

\begin{figure*}[t]
\hspace*{-0.5cm}
\includegraphics[width=18.7 cm,height=8.3cm ]{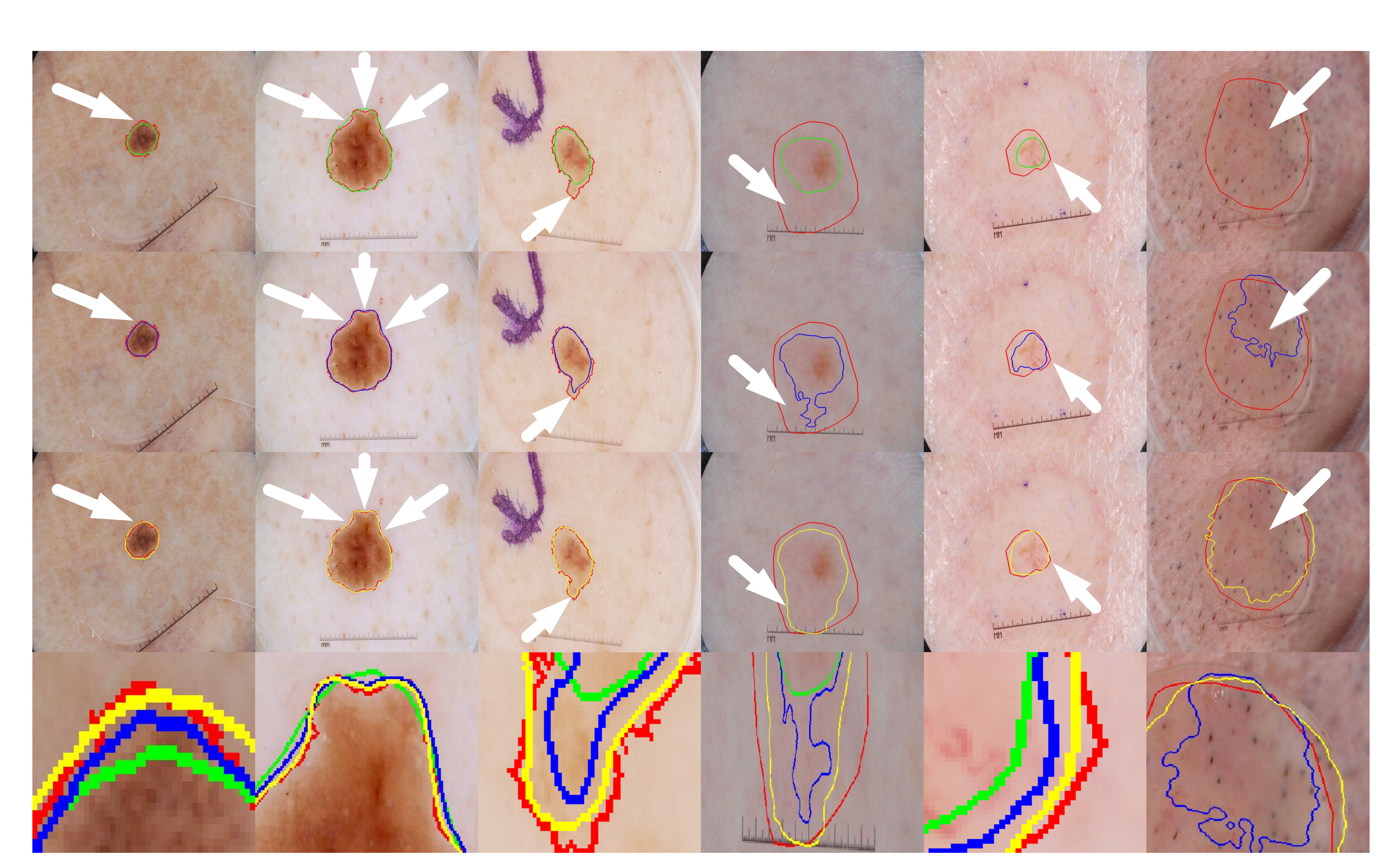}
\vspace*{-0.5cm}
\caption{Qualitative segmented result comparison on ISBI 2017 database. First row to third row: segmentation results by baseline (green lines), baseline +biDFL (blue lines) and baseline+biDFL+mCDF (yellow lines) respectively, where ground truths are annotated as red lines. Last row: detailed segmentation results of sub-regions (marked by white arrows in the corresponding columns).}
\label{fig:5}
\end{figure*}

We also qualitatively analyze the effectiveness of the proposed architecture on skin lesion prediction. Figs. \ref{fig:4} and \ref{fig:5} show experimental results on some challenging dermoscopic images, i.e. skin lesions with variable scale, low contrast, artifact, as well as irregular convex and concave boundary. Compared to the segmentation results of baseline architecture, our proposed network achieves more precise prediction for complex skin lesion boundary. For example, images in the first column of Fig. \ref{fig:5} show the segmentation results on skin lesion with convex boundary. The proposed model has the capability to make the segmentation result converge around the convex boundary, while baseline model results in the large interval zone between segmentation and ground truth. For skin lesion with concave boundary (e.g. images in the fourth column of Fig. \ref{fig:4}), baseline model just yields a smooth boundary which loses parts of the geometrical structure of the skin lesion. By contrast, our proposed model achieves more detailed interpretation of skin lesion with concave boundary. In addition, images in the last column of Fig. \ref{fig:4} show skin lesion with low contrast that is difficult for baseline model to generate discriminative feature maps and make a correct prediction. Our proposed model alleviates the limitation of the baseline model on low contrast lesion segmentation and produces a better delineation of lesion boundary. Moreover, there is an interesting observation from Fig. \ref{fig:5} (i.e. images in the sixth column) that the segmented results by baseline with biDFL generate some holes in the skin lesion prediction map, while the proposed multi-scale consistent decision fusion can effectively refine the false inconsistent prediction and improve overall segmentation result. Qualitative analyses of the experimental results in Figs. \ref{fig:4} and \ref{fig:5} clearly exemplify the validity of the proposed model on skin lesion segmentation.

\begin{table*}
\centering
\hspace*{0.6cm}
\setlength{\abovecaptionskip}{0pt}
\caption{Performance comparison of different feature passing architectures (mean$\pm$standard deviation,\%)}
\renewcommand\arraystretch{1.3}
\setlength{\tabcolsep}{4pt}
\label{tab:aspp}
\begin{tabular}{lllllllll}
\toprule
\toprule
\multicolumn{1}{l}{\multirow{2}{*}{Method}} &\multicolumn{4}{c}{ISBI 2017} \\
\cline{2-5}
 &\multicolumn{1}{c}{JA} &\multicolumn{1}{c}{DI} &\multicolumn{1}{c}{AC} &\multicolumn{1}{c}{GM} \\
\midrule
Baseline &75.74$\pm$20.91 &84.10$\pm$18.93 &93.13$\pm$14.02 &88.47$\pm$17.88\\
Baseline+ASPP &76.90$\pm$20.51 &84.96$\pm$18.10 &93.46$\pm$10.42 &89.40$\pm$16.17\\
Baseline+DenseASPP &77.32$\pm$19.99 &85.32$\pm$16.85 &93.53$\pm$11.57 &88.97$\pm$16.08\\
Baseline+biDFL &78.87$\pm$19.15 &86.51$\pm$16.01 &94.01$\pm$10.68 &89.73$\pm$15.43\\
\bottomrule
\bottomrule
\end{tabular}
\end{table*}

\subsubsection{Evaluation of detailed branch of the proposed feature learning scheme} To investigate the effect of the proposed biDFL strategy on skin lesion segmentation, we conduct several controlled experiments with variable conditions for feature learning scheme. Table \ref{tab:2} shows the performance of dermoscopic feature learning with message propagation in single forward and backward directions (denoted by ${\rm DFL}_{+}$ and ${\rm DFL}_{-}$, respectively). 
Compared to the baseline model, the proposed framework with feature passing along single direction (forward or backward) improves the JA by 1.5\%-2.6\%, which validates the effectiveness of directional feature propagation on skin lesion segmentation. Moreover, the combination of feature passing along two complementary directions further boosts the high-level parsing performance of the network, resulting in an additional increase of the JA by 0.5\%-1.2\%. We also observe that feature passing along the forward direction performs slightly better than feature passing along the backward direction on lesion detection, i.e. improving the JA by 0.1\%-0.3\%.

Furthermore, we also compare the proposed framework with the atrous spatial pyramid pooling (ASPP)~\cite{chen2018deeplab} and DenseASPP~\cite{yang2018denseaspp} in Table \ref{tab:aspp}. ASPP directly concatenates multiple atrous-convolved features with different dilation rates into a final feature representation. DenseASPP emphasizes generating features that cover a large scale range in a dense way. It achieves multi-scale feature representation by stacking a set of atrous convolutional layers. The feature maps generated at the smallest scale dominate the feature representation in DenseASPP, since they participate in producing feature maps at each scale. In contrast to ASPP and DenseASPP feature concatenation schemes, our method focuses on learning relationship between different parts of lesion anatomical structure by passing messages among different receptive fields within two complementary directions, thus local and contextual features cooperate with each other effectively to improve the feature discriminative representation. The results in Table \ref{tab:aspp} show that the proposed feature learning framework outperforms the ASPP and DenseASPP, i.e. improving the JA by 1.6\%-2.0\%.

\begin{table}
\centering
\hspace*{0.7cm}
\setlength{\abovecaptionskip}{0pt}
\caption{Performance comparison of different local region sizes (mean$\pm$standard deviation,\%)}
\renewcommand\arraystretch{1.3}
\setlength{\tabcolsep}{4pt}
\label{tab:3r}
\begin{tabular}{clllll}
\toprule
\toprule
\multicolumn{1}{c}{Size} &\multicolumn{1}{c}{JA} &\multicolumn{1}{c}{DI} &\multicolumn{1}{c}{AC} &\multicolumn{1}{c}{GM}\\
\midrule
$3\times3$ &80.52$\pm$17.79 &87.73$\pm$14.91 &94.37$\pm$9.79 &91.05$\pm$13.38 \\
$5\times5$ &79.97$\pm$18.28 &87.33$\pm$15.42 &94.14$\pm$10.13 &90.39$\pm$14.18 \\
ours  &81.47$\pm$16.69 &88.54$\pm$13.80 &94.65$\pm$9.64 &92.67$\pm$13.22  \\
\bottomrule
\bottomrule
\end{tabular}
\end{table}

\subsubsection{Evaluation of different local region sizes of multiscale consistent decision fusion}


We investigate the lesion segmented performance of local region $L_{p}$ with different size $l\times l$ for multiscale consistent decision fusion, as shown in Table \ref{tab:3r}. If the local region size is fixed, the performance of a small size $3\times3$ is slightly better than that of a large one $5\times5$. However, in our method, the size $l\times l$ that is progressively increased for different layers, produces visible better segmentation performance, i.e. enhancing the JA by 1.0\%-1.5\%. 


\subsubsection{Evaluation of different number of training images}

The ground truths for skin lesion segmentation are taken from International Skin Imaging Collaboration (ISIC), where they are created by an expert clinician. Though human is not error-free, ISIC annotations of skin lesions are quite reliable for performance assessment. Thus we evaluate the performance of our proposed approach in the experiment based on the human annotated ground truths. To analyze the effect of overfitting, we train the model with different number of training images on ISBI 2017 database. Specifically, we randomly select 500, 1000, and 1500 training images for network training. Table. \ref{tab:30} shows the segmentation performance of different number of training samples. With the increase of the number of training images, the segmentation performance JA is progressively improved, since more general information (less overfitting to specific training data) can be learned from training samples with growing number and diversity.

\subsubsection{Evaluation of the sensitivity of dilated rates}

To investigate how different settings of the dilated rates affect the segmentation performance, we conduct experiments with different dilated rate sets on ISBI 2017 database, as shown in Table \ref{tab:30r}. Compared to the dilation rate set (3, 6, 12, 18), the one used in ASPP~\cite{chen2018deeplab} (6, 12, 18, 24) produces better segmentation performance, i.e. improving the JA by 0.3\%. The dilated rate set designed in our work further enhances the lesion segmentation performance by 0.6\% in JA.

\begin{table}[t]
\centering
\hspace*{0.7cm}
\setlength{\abovecaptionskip}{0pt}
\caption{Performance comparison of different number of training samples (mean$\pm$standard deviation,\%)}
\renewcommand\arraystretch{1.3}
\setlength{\tabcolsep}{3.5pt}
\label{tab:30}
\begin{tabular}{cllll}
\toprule
\toprule
\multicolumn{1}{l}{\multirow{1}{*}{Number}} &\multicolumn{1}{c}{JA} &\multicolumn{1}{c}{DI} &\multicolumn{1}{c}{AC} &\multicolumn{1}{c}{GM}\\
\midrule
500 &76.70$\pm$20.51 &84.73$\pm$18.29 &92.95$\pm$12.69 &88.49$\pm$16.92\\
1000 &78.89$\pm$19.15 &86.47$\pm$16.41 &93.53$\pm$11.58 &89.52$\pm$14.61 \\
1500  &80.31$\pm$18.32 &87.55$\pm$15.34 &93.93$\pm$11.17 &90.26$\pm$14.85 \\
2000 &81.47$\pm$16.69 &88.54$\pm$13.80 &94.65$\pm$9.64 &92.67$\pm$13.22 \\
\bottomrule
\bottomrule
\end{tabular}
\end{table}

\begin{table}
\centering
\hspace*{0.7cm}
\setlength{\abovecaptionskip}{0pt}
\caption{Performance comparison of different dilated rate sets (mean$\pm$standard deviation,\%)}
\renewcommand\arraystretch{1.3}
\setlength{\tabcolsep}{2pt}
\label{tab:30r}
\begin{tabular}{clllll}
\toprule
\toprule
\multicolumn{1}{c}{Size} &\multicolumn{1}{c}{JA} &\multicolumn{1}{c}{DI} &\multicolumn{1}{c}{AC} &\multicolumn{1}{c}{GM}\\
\midrule
$(3,6,12,18)$ &80.64$\pm$17.23 &87.94$\pm$14.32 &94.31$\pm$10.28 &90.87$\pm$13.98 \\
$(6,12,18,24)$ &80.89$\pm$17.03 &88.11$\pm$14.12 &94.47$\pm$9.65 &91.03$\pm$13.90 \\
$(3,6,12,18,24)$ &81.47$\pm$16.69 &88.54$\pm$13.80 &94.65$\pm$9.64 &92.67$\pm$13.22  \\
\bottomrule
\bottomrule
\end{tabular}
\end{table}


\subsection{Comparison to Other Published Methods}

\begin{figure*}[t]
\includegraphics[width=18 cm,height=6.8cm ]{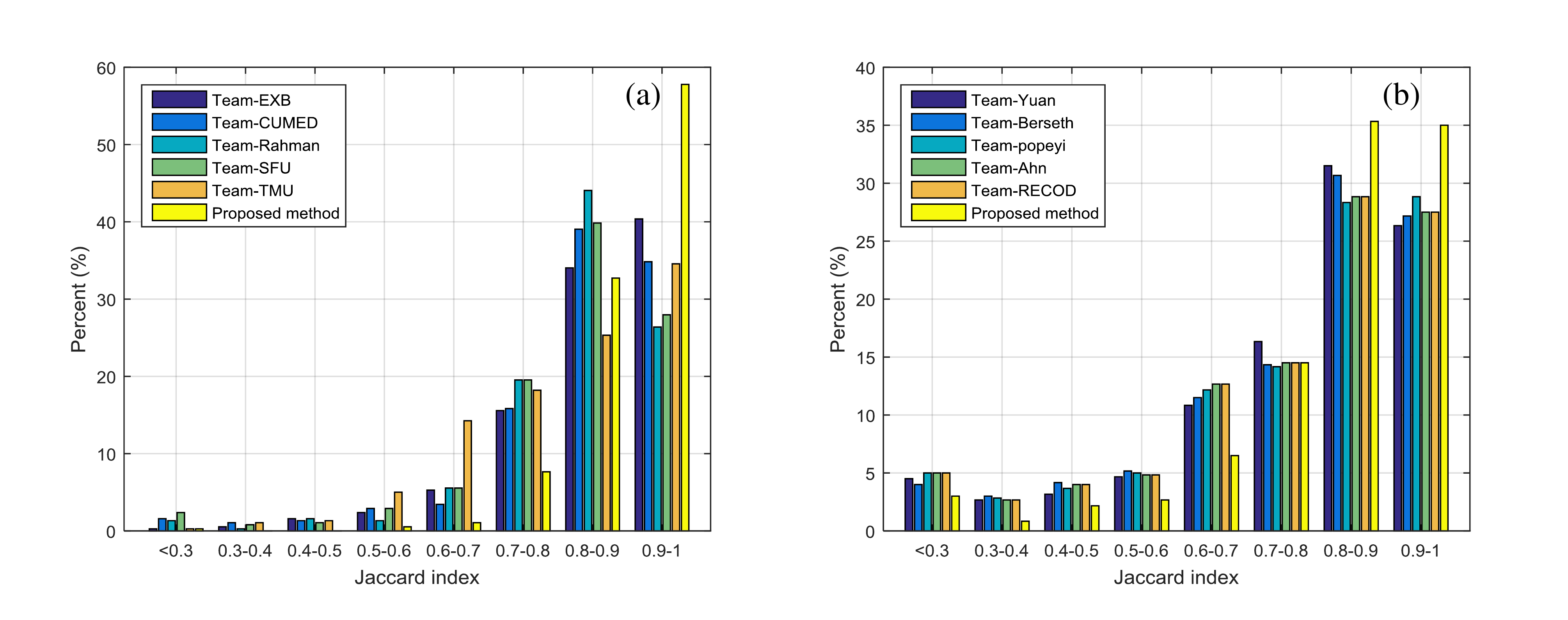}
\vspace*{-0.5cm}
\caption{Distribution comparison in terms of Jaccard index. (a) ISBI 2016 database. (b) ISBI 2017 database.}
\label{fig:6}
\end{figure*}

\begin{table}[t]
\centering
\hspace*{0.7cm}
\setlength{\abovecaptionskip}{0pt}
\caption{Performance comparison with other methods on ISBI 2016 database (\%)}
\renewcommand\arraystretch{1.2}
\setlength{\tabcolsep}{9pt}
\label{tab:4}
\begin{tabular}{llll}
\toprule
\toprule
\multicolumn{1}{l}{\multirow{1}{*}{Method}} &\multicolumn{1}{c}{JA} &\multicolumn{1}{c}{DI} &\multicolumn{1}{c}{AC} \\
\midrule
Team-EXB~\cite{timmurphy16} &84.30 &91.00  &95.30 \\
Team-CUMED~\cite{yu2017automated} &82.90 &89.70 &94.90 \\
Team-Rahman~\cite{timmurphy16} &82.20 &89.50  &95.20 \\
Team-SFU~\cite{timmurphy16} &81.10 &88.50 &94.40 \\
Team-TMU~\cite{timmurphy16} &81.00 &88.80 &94.60 \\
Arroyo and Zapirain~\cite{garcia2018segmentation} & 79.10 &86.90 &93.40 \\
Yuan \emph {et al}.~\cite{yuan2017automatic} &84.70 &91.2 &95.50 \\
Bi \emph {et al}.~\cite{bi2017dermoscopic} &84.64 &91.18 &95.51 \\
Yuan \emph {et al}.~\cite{yuan2017improving}&84.90 &91.30 &95.70 \\
Bi \emph {et al}.~\cite{bi2019step} &85.92 &91.77 &95.78 \\
\textbf{Proposed method}  &\textbf{88.12} &\textbf{93.33} &\textbf{96.75}  \\
\bottomrule
\bottomrule
\end{tabular}
\end{table}

\begin{table}
\centering
\hspace*{0.7cm}
\setlength{\abovecaptionskip}{0pt}
\caption{Performance comparison with other methods on ISBI 2017 database (\%)}
\renewcommand\arraystretch{1.2}
\setlength{\tabcolsep}{9pt}
\label{tab:5}
\begin{tabular}{llll}
\toprule
\toprule
\multicolumn{1}{l}{\multirow{1}{*}{Method}} &\multicolumn{1}{c}{JA} &\multicolumn{1}{c}{DI} &\multicolumn{1}{c}{AC} \\
\midrule
Team-Yuan~\cite{yuan2017improving} &76.50 &84.90 &93.40  \\
Team-Berseth~\cite{timmurphy17} &76.20 &84.70 &93.20 \\
Team-popleyi~\cite{timmurphy17} &76.00 &84.40 &93.40  \\
Team-Ahn~\cite{timmurphy17} &75.80 &84.40 &93.40 \\
Team-RECOD~\cite{timmurphy17}  &75.40 &83.90 &93.10 \\
Arroyo and Zapirain~\cite{garcia2018segmentation} &66.50 &76.00 &88.40 \\
Lin \emph {et al}.~\cite{lin2017skin} &65.00 &79.00 &N.A \\
Al-masni \emph {et al}.~\cite{al2018skin} &77.11 &87.08 &94.03\\
Bi \emph {et al}.~\cite{bi2019step} &77.73 &85.66 &94.08 \\
\textbf{Proposed method}  &\textbf{81.47} &\textbf{88.54} &\textbf{94.65} \\
\bottomrule
\bottomrule
\end{tabular}
\end{table}

We compare our method with the techniques from top 10 different teams in the competitions of challenge of ISBI 2016 
and 2017 
databases and 7 other top published methods. All compared results are taken from their respective publications. Tables \ref{tab:4} and \ref{tab:5} illustrate the segmentation performance on ISBI 2016 and 2017 databases with different methods respectively. Compared to those state-of-the-arts architectures, our proposed model produces the best segmentation performance on skin lesions in ISBI 2016 and 2017 database consistently. For ISBI 2017 database that contains more complex skin lesions difficult to be distinguished from background, our proposed method shows more performance gain on skin lesion segmentation, resulting in the JA improvement over 3.7\% from the second best approach. Tables \ref{tab:6} and \ref{tab:7} list the segmentation performance of different methods on melanoma and non-melanoma cases separately, where the proposed method outperforms other techniques in both cases consistently. Segmentation of melanoma is more difficult than non-melanoma due to severe inhomogeneous of lesion pattern. Higher performance gain of melanoma segmentation than non-melanoma segmentation in Table \ref{tab:7} indicates the effectiveness of the proposed method on melanoma detection, which is beneficial to the further inspection of melanoma.

Fig. \ref{fig:6} shows the distribution comparison in terms of Jaccard index on two skin lesion databases. For ISBI 2016 database, the proposed architecture yields 57.8\% segmentation results whose Jaccard index higher than 90\%, leading to the noticeable improvement than other methods, e.g. 17.4\% higher than the second best Team-EXB~\cite{timmurphy16}. For more difficult ISBI 2017 database, the proposed architecture produces 70.3\% segmentation results with Jaccard index larger than 80\%, increasing 12.6\% compared to Team-Yuan~\cite{yuan2017improving}. Analysis of distribution of Jaccard index indicates the utility and stability of our proposed method on skin lesion segmentation, where segmentation results with high Jaccard index occupy the majority of both ISBI 2016 and 2017 databases, significantly better than other methods used for comparison.

\begin{table}[t]
\centering
\setlength{\abovecaptionskip}{0pt}
\setlength{\belowcaptionskip}{100pt}
\caption{Performance comparison with other methods on ISBI 2016 database for melanoma and non-melanoma cases (\%)}
\renewcommand\arraystretch{1.2}
\setlength{\tabcolsep}{7pt}
\label{tab:6}
\begin{tabular}{lcccc}
\toprule
\toprule
\multicolumn{1}{l}{\multirow{2}{*}{Method}} &\multicolumn{2}{c}{Melanoma} &\multicolumn{2}{c}{Non-Melanoma}\\
\cline{2-3} \cline{4-5}
&JA &DI  &JA &DI \\
\midrule
Team-EXB~\cite{timmurphy16} &82.94 &90.11 &84.64 &91.18\\
Team-CUMED~\cite{yu2017automated} &82.90 &89.98 &82.95 &89.68  \\
Team-Rahman~\cite{timmurphy16} &82.65 &89.93 &82.04 &89.44 \\
Team-SFU~\cite{timmurphy16} &81.88 &89.44 &80.88 &88.32\\
Team-TMU~\cite{timmurphy16}  &82.31 &89.68 &80.73 &88.58  \\
Arroyo and Zapirain~\cite{garcia2018segmentation} &80.94 &88.60 &86.51 &78.68 \\
Bi \emph {et al}.~\cite{bi2017dermoscopic} &85.84 &92.03 &84.34 &90.97 \\
Bi \emph {et al}.~\cite{bi2019step} &85.62 &91.72 &85.60 &91.78 \\
\textbf{Proposed method}  &\textbf{88.38} &\textbf{93.62} &\textbf{88.06} &\textbf{93.26} \\
\bottomrule
\bottomrule
\end{tabular}
\end{table}

\begin{table}
\centering
\setlength{\abovecaptionskip}{0pt}
\setlength{\belowcaptionskip}{100pt}
\caption{Performance comparison with other methods on ISBI 2017 database for melanoma and non-melanoma cases (\%)}
\renewcommand\arraystretch{1.2}
\setlength{\tabcolsep}{7pt}
\label{tab:7}
\begin{tabular}{lcccc}
\toprule
\toprule
\multicolumn{1}{l}{\multirow{2}{*}{Method}} &\multicolumn{2}{c}{Melanoma} &\multicolumn{2}{c}{Non-Melanoma}\\
\cline{2-3} \cline{4-5}
&JA &DI  &JA &DI \\
\midrule
Team-Yuan~\cite{yuan2017improving} &71.20 &81.04 &77.78 &85.81\\
Team-Berseth~\cite{timmurphy17} &68.82 &79.07 &78.02 &86.07  \\
Team-popleyi~\cite{timmurphy17} &69.28 &79.62 &77.60 &85.50 \\
Team-Ahn~\cite{timmurphy17} &69.06 &79.48 &77.45 &85.38\\
Team-RECOD~\cite{timmurphy17}  &68.78 &79.08 &77.00 &85.11  \\
Arroyo and Zapirain~\cite{garcia2018segmentation} &65.81 &76.56 &66.70 &75.88 \\
Bi \emph {et al}.~\cite{bi2019step} &72.18 &81.65 &79.07 &86.63 \\
\textbf{Proposed method}  &\textbf{77.26} &\textbf{85.35} &\textbf{82.49} &\textbf{89.31}  \\
\bottomrule
\bottomrule
\end{tabular}
\end{table}

\subsection{Result summary}

The proposed network achieves state-of-the-art segmentation performance on two publicly available skin lesion databases and in both melanoma and non-melanoma cases consistently. Results recorded in Table \ref{tab:1} and illustrated in Figs. \ref{fig:4} and \ref{fig:5}, show the significant performance gain brought by the two proposed techniques, biDFL and mCDF. This superiority of the proposed biDFL comes from the informative feature passing through two complementary directions, which makes the feature maps receive discriminative information from complex spatial configuration of the skin lesion. The proposed mCDF selectively fuses decision scores of multi-scale features by checking their consistency in a spatial local area. Extensive comparisons with other reported methods have shown that our approach consistently performs better than others on skin lesion segmentation (see Tables \ref{tab:4}-\ref{tab:7}). We attribute this profit to the fact that we investigate more insightful relationship between skin lesions and their informative context, as well as the consistency of the decision from multiple classification layers, which have not yet been well explored by the previous studies. Moreover, in addition to the task of skin lesion segmentation, the proposed architecture is flexible to be extended to other field of image analysis.
\begin{figure}[t]
\hspace*{-0.3cm}
\vspace*{-0.3cm}
\includegraphics[width=9.3 cm,height=4cm ]{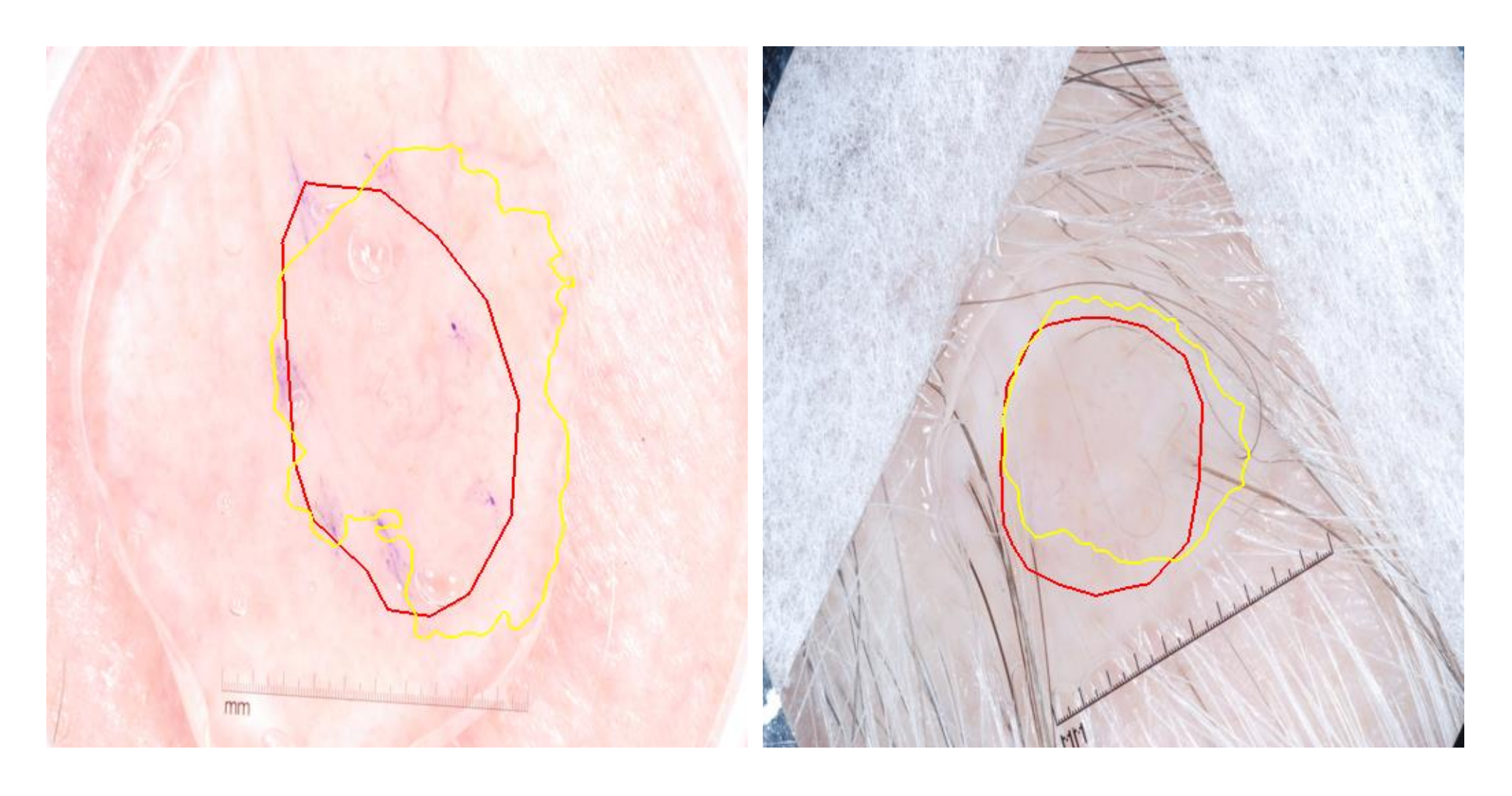}
\caption{Examples of some cases that need further improvement. Red and yellow lines represent ground truths and segmentation results.}
\label{fig:7}
\end{figure}


Although our proposed network has generated high performance for lesion segmentation, it should be noted that there are some lesion segmentation cases that can be further improved, as shown in Fig. \ref{fig:7}. Most of those skin lesions are of low contrast and have irregular structure. The main reason for insufficient segmentation of those skin lesions is the scarcity of the relative dermoscopic images in the training data. One way to further improve the segmentation performance is to learn productive feature representation from more accessible training samples. Images with skin lesion acquired from mobile computing devices such as smartphones will provide an appealing way for efficiently collecting lesion images and self-monitoring of melanoma. It is an interesting project worthy of further investigation.

\section{Conclusion}

In this paper, we propose a bi-directional dermoscopic feature learning framework to generate a substantially rich description for skin lesion structure. Feature information passing through two complementary directions among high-level layers gives a significant improvement of the parsing ability of the network. Furthermore, we propose a multi-scale consistent decision fusion to selectively focus on more consistent decisions generated from multiple classification layers, which achieves more reliable prediction for skin lesion on dermoscopic images. Both qualitative and quantitative analyses of segmentation performance on two publicly available databases show the superiority of the proposed method, especially for tiny skin lesion structure delineation and complex boundary localization. As an effective and efficient segmentation tool, the proposed network is flexible to be extended to solve many other image segmentation problems.

\section*{Acknowledgment}

The authors would like to thank the organizers of \textit{International Symposium on Biomedical Imaing 2016 and 2017} for kindly providing benchmark databases and annotations.

\ifCLASSOPTIONcaptionsoff
  \newpage
\fi

\bibliographystyle{IEEEbib}
\bibliography{Refskin}

\begin{thebibliography}{10}

\bibitem{ma2013analysis}
L.~Ma and R.~C. Staunton,
\newblock ``Analysis of the contour structural irregularity of skin lesions
  using wavelet decomposition,''
\newblock {\em Pattern Recogn.}, vol. 46, no. 1, pp. 98--106, 2013.

\bibitem{garcia2018segmentation}
J.~Garcia-Arroyo and B.~Garcia-Zapirain,
\newblock ``Segmentation of skin lesions in dermoscopy images using fuzzy
  classification of pixels and histogram thresholding,''
\newblock {\em Comput. Meth. Prog. Bio.}, vol. 168, pp. 11--19, 2019.

\bibitem{barata2017development}
C.~Barata, M.~E. Celebi, and J.~S. Marques,
\newblock ``Development of a clinically oriented system for melanoma
  diagnosis,''
\newblock {\em Pattern Recogn.}, vol. 69, pp. 270--285, 2017.

\bibitem{xie2017melanoma}
F.~Xie, H.~Fan, Y.~Li, Z.~Jiang, R.~Meng, and A.~Bovik,
\newblock ``Melanoma classification on dermoscopy images using a neural network
  ensemble model,''
\newblock {\em IEEE Trans. Med. Imaging}, vol. 36, no. 3, pp. 849--858, 2017.

\bibitem{kharazmi2017automated}
P.~Kharazmi, M.~I. AlJasser, H.~Lui, Z.~J. Wang, and T.~K. Lee,
\newblock ``Automated detection and segmentation of vascular structures of skin
  lesions seen in dermoscopy, with an application to basal cell carcinoma
  classification,''
\newblock {\em IEEE J. Biomed. Health Inform}, vol. 21, no. 6, pp. 1675--1684,
  2017.

\bibitem{shimizu2015four}
K.~Shimizu, H.~Iyatomi, M.~Celebi, K.~Norton, and M.~Tanaka,
\newblock ``Four-class classification of skin lesions with task decomposition
  strategy,''
\newblock {\em IEEE Trans. Biomed. Eng}, vol. 62, no. 1, pp. 274--283, 2015.

\bibitem{yu2017automated}
L.~Yu, H.~Chen, Q.~Dou, J.~Qin, and P.~Heng,
\newblock ``Automated melanoma recognition in dermoscopy images via very deep
  residual networks,''
\newblock {\em IEEE Trans. Med. Imaging}, vol. 36, no. 4, pp. 994--1004, 2017.

\bibitem{cavalcanti2011automated}
P.~Cavalcanti and J.~Scharcanski,
\newblock ``Automated prescreening of pigmented skin lesions using standard
  cameras,''
\newblock {\em Comput. Med. Imag. Grap.}, vol. 35, no. 6, pp. 481--491, 2011.

\bibitem{ahn2017saliency}
E.~Ahn, J.~Kim, L.~Bi, A.~Kumar, C.~Li, M.~Fulham, and D.~D. Feng,
\newblock ``Saliency-based lesion segmentation via background detection in
  dermoscopic images,''
\newblock {\em IEEE J. Biomed. Health Inform}, vol. 21, no. 6, pp. 1685--1693,
  2017.

\bibitem{diaz2018dermaknet}
I.~Diaz,
\newblock ``Dermaknet: Incorporating the knowledge of dermatologists to
  convolutional neural networks for skin lesion diagnosis,''
\newblock {\em IEEE J. Biomed. Health Inform}, 2018.

\bibitem{yuan2017automatic}
Y.~Yuan, M.~Chao, and Y.~Lo,
\newblock ``Automatic skin lesion segmentation using deep fully convolutional
  networks with jaccard distance,''
\newblock {\em IEEE Trans. Med. Imaging}, vol. 36, no. 9, pp. 1876--1886, 2017.

\bibitem{bi2017dermoscopic}
L.~Bi, J.~Kim, E.~Ahn, A.~Kumar, M.~Fulham, and D.~Feng,
\newblock ``Dermoscopic image segmentation via multi-stage fully convolutional
  networks,''
\newblock {\em IEEE Trans. Biomed. Eng}, vol. 64, no. 9, pp. 2065--2074, 2017.

\bibitem{bi2019step}
L.~Bi, J.~Kim, E.~Ahn, A.~Kumar, D.~Feng, and M.~Fulham,
\newblock ``Step-wise integration of deep class-specific learning for
  dermoscopic image segmentation,''
\newblock {\em Pattern Recogn.}, vol. 85, pp. 78--89, 2019.

\bibitem{chen2018deeplab}
L.~Chen, G.~Papandreou, I.~Kokkinos, K.~Murphy, and A.~Yuille,
\newblock ``Deeplab: Semantic image segmentation with deep convolutional nets,
  atrous convolution, and fully connected crfs,''
\newblock {\em IEEE Trans. Pattern Anal. Mach. Intell.}, vol. 40, no. 4, pp.
  834--848, 2018.

\bibitem{yang2018denseaspp}
M.~Yang, K.~Yu, C.~Zhang, Z.~Li, and K.~Yang,
\newblock ``Denseaspp for semantic segmentation in street scenes,''
\newblock in {\em Proceedings of the IEEE Conference on Computer Vision and
  Pattern Recognition}, 2018, pp. 3684--3692.

\bibitem{khandpur2012skin}
S.~Khandpur and M.~Ramam,
\newblock ``Skin tumours,''
\newblock {\em Journal of cutaneous and aesthetic surgery}, vol. 5, no. 3, pp.
  159, 2012.

\bibitem{long2015fully}
J.~Long, E.~Shelhamer, and T.~Darrell,
\newblock ``Fully convolutional networks for semantic segmentation,''
\newblock in {\em Proceedings of the IEEE conference on computer vision and
  pattern recognition}, 2015, pp. 3431--3440.

\bibitem{shuai2018scene}
B.~Shuai, Z.~Zuo, B.~Wang, and G.~Wang,
\newblock ``Scene segmentation with dag-recurrent neural networks,''
\newblock {\em IEEE Trans. Pattern Anal. Mach.Intell.}, vol. 40, no. 6, pp.
  1480--1493, 2018.

\bibitem{ding2020semantictip}
H.~Ding, X.~Jiang, B.~Shuai, A.~Q. Liu, and G.~Wang,
\newblock ``Semantic segmentation with context encoding and multi-path
  decoding,''
\newblock {\em IEEE Transactions on Image Processing}, 2020.

\bibitem{shuai2019toward}
B.~Shuai, H.~Ding, T.~Liu, G.~Wang, and X.~Jiang,
\newblock ``Toward achieving robust low-level and high-level scene parsing,''
\newblock {\em IEEE Trans. Image Process.}, vol. 28, no. 3, pp. 1378--1390,
  2019.

\bibitem{ding2019boundary}
H.~Ding, X.~Jiang, A.~Q. Liu, N.~M. Thalmann, and G.~Wang,
\newblock ``Boundary-aware feature propagation for scene segmentation,''
\newblock in {\em Proceedings of the IEEE International Conference on Computer
  Vision}, 2019, pp. 6819--6829.

\bibitem{wang2019dermoscopic}
X.~Wang, H.~Ding, and X.~Jiang,
\newblock ``Dermoscopic image segmentation through the enhanced high-level
  parsing and class weighted loss,''
\newblock in {\em Proceedings of the IEEE International Conference on Image
  Processing}, 2019, pp. 245--249.

\bibitem{yuksel2009accurate}
M.~Y{\"u}ksel and M.~Borlu,
\newblock ``Accurate segmentation of dermoscopic images by image thresholding
  based on type-2 fuzzy logic,''
\newblock {\em IEEE Trans. Fuzzy Syst.}, vol. 17, no. 4, pp. 976--982, 2009.

\bibitem{emre2013lesion}
M.~Celebi, Q.~Wen, S.~Hwang, H.~Iyatomi, and G.~Schaefer,
\newblock ``Lesion border detection in dermoscopy images using ensembles of
  thresholding methods,''
\newblock {\em Skin Research and Technology}, vol. 19, no. 1, pp. e252--e258,
  2013.

\bibitem{schmid1999segmentation}
P.~Schmid,
\newblock ``Segmentation of digitized dermatoscopic images by two-dimensional
  color clustering,''
\newblock {\em IEEE Trans. Med. Imaging}, vol. 18, no. 2, pp. 164--171, 1999.

\bibitem{zhou2009anisotropic}
H.~Zhou, G.~Schaefer, A.~Sadka, and M.~Celebi,
\newblock ``Anisotropic mean shift based fuzzy c-means segmentation of
  deroscopy images,''
\newblock {\em IEEE J. Sel. Top. Signal Process.}, vol. 3, no. 1, pp. 26--34,
  2009.

\bibitem{zhou2011gradient}
H.~Zhou, G.~Schaefer, M.~Celebi, F.~Lin, and T.~Liu,
\newblock ``Gradient vector flow with mean shift for skin lesion
  segmentation,''
\newblock {\em Comput. Med. Imag. Grap.}, vol. 35, no. 2, pp. 121--127, 2011.

\bibitem{ma2016novel}
Z.~Ma and J.~M.~R. Tavares,
\newblock ``A novel approach to segment skin lesions in dermoscopic images
  based on a deformable model,''
\newblock {\em IEEE J. Biomed. Health Inform}, vol. 20, no. 2, pp. 615--623,
  2016.

\bibitem{wang2011modified}
H.~Wang, R.~Moss, X.~Chen, R.~Stanley, W.~Stoecker, M.~Celebi, J.~Malters,
  J.~Grichnik, A.~Marghoob, H.~Rabinovitz, et~al.,
\newblock ``Modified watershed technique and post-processing for segmentation
  of skin lesions in dermoscopy images,''
\newblock {\em Comput. Med. Imag. Grap.}, vol. 35, no. 2, pp. 116--120, 2011.

\bibitem{he2012automatic}
Y.~He and F.~Xie,
\newblock ``Automatic skin lesion segmentation based on texture analysis and
  supervised learning,''
\newblock in {\em Asian Conference on Computer Vision}. Springer, 2012, pp.
  330--341.

\bibitem{jahanifar2018supervised}
M.~Jahanifar, N.~Z. Tajeddin, B.~M. Asl, and A.~Gooya,
\newblock ``Supervised saliency map driven segmentation of lesions in
  dermoscopic images,''
\newblock {\em IEEE J. Biomed. Health Inform}, 2018.

\bibitem{mei2019deepdeblur}
J.~Mei, Z.~Wu, X.~Chen, Y.~Qiao, H.~Ding, and X.~Jiang,
\newblock ``Deepdeblur: text image recovery from blur to sharp,''
\newblock {\em Multimedia Tools and Applications}, vol. 78, no. 13, pp.
  18869--18885, 2019.

\bibitem{liu2019feature}
J.~Liu, H.~Ding, A.~Shahroudy, L.~Duan, X.~Jiang, G.~Wang, and A.~K. Chichung,
\newblock ``Feature boosting network for 3d pose estimation,''
\newblock {\em IEEE transactions on pattern analysis and machine intelligence},
  2019.

\bibitem{pereira2016brain}
S.~Pereira, A.~Pinto, V.~Alves, and C.~Silva,
\newblock ``Brain tumor segmentation using convolutional neural networks in mri
  images,''
\newblock {\em IEEE Trans. Med. Imaging}, vol. 35, no. 5, pp. 1240--1251, 2016.

\bibitem{wang2018interactive}
G.~Wang, W.~Li, M.~Zuluaga, R.~Pratt, P.~Patel, M.~Aertsen, T.~Doel, A.~David,
  J.~Deprest, S.~Ourselin, et~al.,
\newblock ``Interactive medical image segmentation using deep learning with
  image-specific fine-tuning,''
\newblock {\em IEEE Trans. Med. Imaging}, 2018.

\bibitem{chen2016dcan}
H.~Chen, X.~Qi, L.~Yu, and P.~Heng,
\newblock ``Dcan: deep contour-aware networks for accurate gland
  segmentation,''
\newblock in {\em Proceedings of the IEEE conference on Computer Vision and
  Pattern Recognition}, 2016, pp. 2487--2496.

\bibitem{setio2016pulmonary}
A.~Setio, F.~Ciompi, G.~Litjens, P.~Gerke, C.~Jacobs, S.~Van~Riel, M.~Wille,
  M.~Naqibullah, C.~S{\'a}nchez, and B.~van Ginneken,
\newblock ``Pulmonary nodule detection in ct images: false positive reduction
  using multi-view convolutional networks,''
\newblock {\em IEEE Trans. Med. Imaging}, vol. 35, no. 5, pp. 1160--1169, 2016.

\bibitem{yan2016multi}
Z.~Yan, Y.~Zhan, Z.~Peng, S.~Liao, Y.~Shinagawa, S.~Zhang, D.~Metaxas, and
  X.~Zhou,
\newblock ``Multi-instance deep learning: Discover discriminative local
  anatomies for bodypart recognition,''
\newblock {\em IEEE Trans. Med. Imaging}, vol. 35, no. 5, pp. 1332--1343, 2016.

\bibitem{gu2019net}
Z.~Gu, J.~Cheng, H.~Fu, K.~Zhou, H.~Hao, Y.~Zhao, T.~Zhang, S.~Gao, and J.~Liu,
\newblock ``Ce-net: Context encoder network for 2d medical image
  segmentation,''
\newblock {\em IEEE Trans. Med. Imaging}, 2019.

\bibitem{zhang2019net}
Z.~Zhang, H.~Fu, H.~Dai, J.~Shen, Y.~Pang, and L.~Shao,
\newblock ``Et-net: A generic edge-attention guidance network for medical image
  segmentation,''
\newblock {\em Proceedings of the International Conference on Medical Image
  Computing and Computer Assisted Intervention Society}, 2019.

\bibitem{woo2018cbam}
S.~Woo, J.~Park, J.~Lee, and I.~So~Kweon,
\newblock ``Cbam: Convolutional block attention module,''
\newblock in {\em Proceedings of the European Conference on Computer Vision},
  2018, pp. 3--19.

\bibitem{hu2018squeeze}
J.~Hu, L.~Shen, and G.~Sun,
\newblock ``Squeeze-and-excitation networks,''
\newblock in {\em Proceedings of the IEEE conference on computer vision and
  pattern recognition}, 2018, pp. 7132--7141.

\bibitem{schlemper2019attention}
J.~Schlemper, O.~Oktay, M.~Schaap, M.~Heinrich, B.~Kainz, B.~Glocker, and
  D.~Rueckert,
\newblock ``Attention gated networks: Learning to leverage salient regions in
  medical images,''
\newblock {\em Medical image analysis}, vol. 53, pp. 197--207, 2019.

\bibitem{wang2019deep}
Y.~Wang, H.~Dou, X.~Hu, L.~Zhu, X.~Yang, M.~Xu, J.~Qin, P.~Heng, T.~Wang, and
  D.~Ni,
\newblock ``Deep attentive features for prostate segmentation in 3d transrectal
  ultrasound,''
\newblock {\em IEEE transactions on medical imaging}, 2019.

\bibitem{yuan2017improving}
Y.~Yuan and Y.~Lo,
\newblock ``Improving dermoscopic image segmentation with enhanced
  convolutional-deconvolutional networks,''
\newblock {\em IEEE J. Biomed. Health Inform}, 2017.

\bibitem{lin2017skin}
B.~Lin, K.~Michael, S.~Kalra, and H.~Tizhoosh,
\newblock ``Skin lesion segmentation: U-nets versus clustering,''
\newblock in {\em Proceedings of the IEEE Symposium Series on Computational
  Intelligence (SSCI)}, 2017, pp. 1--7.

\bibitem{al2018skin}
M.~Al-masni, M.~Al-antari, M.~Choi, S.~Han, and T.~Kim,
\newblock ``Skin lesion segmentation in dermoscopy images via deep full
  resolution convolutional networks,''
\newblock {\em Comput. Meth. Prog. Bio.}, vol. 162, pp. 221--231, 2018.

\bibitem{he2016deep}
K.~He, X.~Zhang, S.~Ren, and J.~Sun,
\newblock ``Deep residual learning for image recognition,''
\newblock in {\em Proceedings of the IEEE conference on computer vision and
  pattern recognition}, 2016, pp. 770--778.

\bibitem{russakovsky2015imagenet}
O.~Russakovsky, J.~Deng, H.~Su, J.~Krause, S.~Satheesh, S.~Ma, Z.~Huang,
  A.~Karpathy, A.~Khosla, M.~Bernstein, et~al.,
\newblock ``Imagenet large scale visual recognition challenge,''
\newblock {\em Int. J. Comput. Vision}, vol. 115, no. 3, pp. 211--252, 2015.

\bibitem{hubel1995eye}
D.~Hubel,
\newblock {\em Eye, brain, and vision.},
\newblock Scientific American Library/Scientific American Books, 1995.

\bibitem{huang2017rapid}
J.~Huang, Y.~Yang, K.~Zhou, X.~Zhao, Q.~Zhou, H.~Zhu, Y.~Yang, C.~Zhang,
  Y.~Zhou, and W.~Zhou,
\newblock ``Rapid processing of a global feature in the on visual pathways of
  behaving monkeys,''
\newblock {\em Front. neurosci.}, vol. 11, pp. 474, 2017.

\bibitem{ahissar2004reverse}
M.~Ahissar and S.l Hochstein,
\newblock ``The reverse hierarchy theory of visual perceptual learning,''
\newblock {\em Trends cogn. sci.}, vol. 8, no. 10, pp. 457--464, 2004.

\bibitem{zoccolan2015invariant}
D.~Zoccolan,
\newblock ``Invariant visual object recognition and shape processing in rats,''
\newblock {\em Behav. brain res.}, vol. 285, pp. 10--33, 2015.

\bibitem{simonyan2014very}
K.~Simonyan and A.~Zisserman,
\newblock ``Very deep convolutional networks for large-scale image
  recognition,''
\newblock {\em arXiv preprint arXiv:1409.1556}, 2014.

\bibitem{peng2017large}
C.~Peng, X.~Zhang, G.~Yu, G.~Luo, and J.~Sun,
\newblock ``Large kernel matters--improve semantic segmentation by global
  convolutional network,''
\newblock in {\em Proceedings of the IEEE conference on computer vision and
  pattern recognition}, 2017, pp. 4353--4361.

\bibitem{timmurphy16}
ISIC 2016,
\newblock ``Skin lesion analysis towards melanoma detection,''
  \url{https://challenge.kitware.com/\#phase/566744dccad3a56fac786787.}

\bibitem{timmurphy17}
ISIC 2017,
\newblock ``Skin lesion analysis towards melanoma detection,''
  \url{https://challenge.kitware.com/#phase/584b0afacad3a51cc66c8e24.}

\bibitem{ding2019semantic}
H.~Ding, X.~Jiang, B.~Shuai, A.~Q. Liu, and G.~Wang,
\newblock ``Semantic correlation promoted shape-variant context for
  segmentation,''
\newblock in {\em Proceedings of the IEEE Conference on Computer Vision and
  Pattern Recognition}, 2019, pp. 8885--8894.

\bibitem{ding2018context}
H.~Ding, X.~Jiang, B.~Shuai, A.~Liu, and G.~Wang,
\newblock ``Context contrasted feature and gated multi-scale aggregation for
  scene segmentation,''
\newblock in {\em Proceedings of the IEEE Conference on Computer Vision and
  Pattern Recognition}, 2018, pp. 2393--2402.

\end{thebibliography}

\end{document}